\PassOptionsToPackage{table}{xcolor}
\documentclass[10pt,twocolumn,letterpaper]{article}
\usepackage[pagenumbers]{cvpr}
\usepackage[T1]{fontenc}

\usepackage{algorithm}
\usepackage{algpseudocode}

\definecolor{cvprblue}{rgb}{0.21,0.49,0.74}
\usepackage[pagebackref,breaklinks,colorlinks,allcolors=cvprblue]{hyperref}

\def\confName{CVPR}
\def\confYear{2027}
\title{EvoKIE: A Co-Evolutionary Framework for Document Key Information Extraction}
\author{%
Shuyi Liu \quad Qi Yao \quad Jun Ma \quad Yanbing Chou\\
Cainiao Group\\
{\tt\small \{susie.lsy, yq223369, jeff.mj, tonychou.zyb\}@cainiao.com}
}
\date{}
\usepackage{array,tabularx,fvextra,textcomp}
\DeclareCaptionFont{casecaption}{\fontsize{8.5}{9.8}\selectfont}
\newcolumntype{P}[1]{>{\raggedright\arraybackslash}p{#1}}
\definecolor{LandmarkRed}{HTML}{D32626}
\definecolor{RegionGreen}{HTML}{16833F}
\definecolor{GTPurple}{HTML}{9228C5}
\definecolor{PredBlue}{HTML}{1662C4}
\newcommand{\dataitem}[1]{\par\noindent\hangindent=0.8em\hangafter=1\makebox[0.8em][l]{\textbullet}#1\par}
\hypersetup{pdftitle={EvoKIE: A Co-Evolutionary Framework for Document Key Information Extraction},pdfauthor={Shuyi Liu, Qi Yao, Jun Ma, Yanbing Chou}}

\begin{document}
\maketitle
\begin{abstract}
Document key information extraction (KIE) at enterprise scale demands accurate and scalable processing across diverse layouts. Many existing methods group documents by layout and reuse group-specific extraction programs to accommodate this diversity while enabling efficient extraction. However, predefined grouping criteria may not support both reliable routing and program reuse, limiting extraction accuracy. We propose EvoKIE, a co-evolutionary KIE framework that jointly refines document grouping, the router, and extraction programs. First, the framework uses routing and extraction failures to refine group boundaries, allowing document grouping to adapt to the performance of the router and extraction programs. Second, an agentic module optimization mechanism refines the router and synthesizes group-specific extraction programs as document grouping evolves. This co-evolution aligns document grouping with the router and extraction programs to improve extraction accuracy. Experiments on three diverse document datasets demonstrate consistent improvements in field-level F1 and substantial gains in document-level accuracy, with improvements of up to 27.67 percentage points over the strongest baselines.
\end{abstract}

\section{Introduction}
\label{sec:introduction}

Document key information extraction (KIE) aims to extract predefined field values from unstructured documents, such as invoices and receipts. Accurate extraction at enterprise scale must accommodate diverse layouts~\cite{wang2023vrdu} and domain-specific field semantics. Multimodal large language models (MLLMs) may struggle to interpret these fields consistently~\cite{shen2026ocrornot}, while supervised fine-tuning requires substantial annotation effort. Nevertheless, enterprise documents often exhibit recurring layouts, with shared field labels and spatial relationships that provide reusable cues for extraction~\cite{gomez2026electricity,zhang2025sail,lin2026twix,bhoi2026cacheed2,nguyen2024vrdsynth}. Many existing methods exploit such regularities with a divide-and-conquer strategy: group documents by layout, synthesize a shared extraction program for each group, and route incoming documents to the appropriate program~\cite{bhoi2026cacheed2,gosmar2026madp,ng2026aibaam}. Document grouping is therefore central to this approach, as it determines both the routing targets and the scope of extraction-program sharing.

\begin{figure*}[t]
    \centering
    \includegraphics[width=\textwidth]{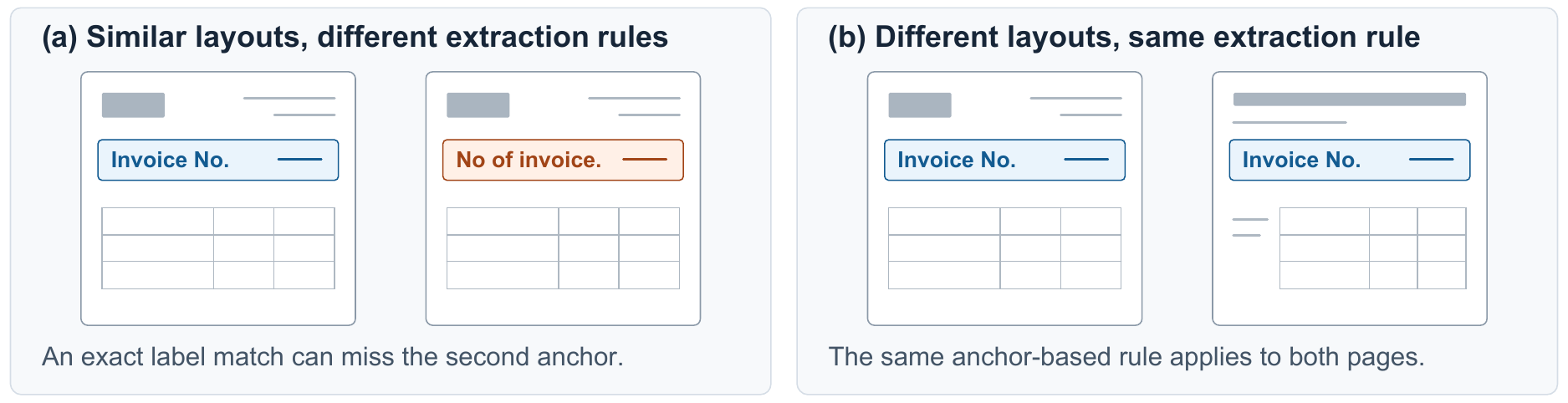}
    \caption{\textbf{Layout similarity does not necessarily imply extraction-rule compatibility.} (a) Two invoices with similar layouts use different labels for the invoice number, causing an exact-match rule targeting ``Invoice No.'' to miss ``No of invoice.'' (b) Despite differences in masthead and table layouts, two invoices share the same invoice-number label and relative value position, allowing a single extraction rule to work for both.}
    \label{fig:layout-rule-compatibility}
\end{figure*}

Determining appropriate document groups is challenging for two reasons. First, grouping must balance routing complexity and extraction-program compatibility~\cite{yan2015hdcnn,sampaio2025templateclustering,bhattacharyya2025blockie}. Merging groups reduces the number of routing targets but may introduce conflicting extraction requirements, whereas splitting groups may improve program compatibility at the cost of finer-grained routing decisions.

Second, document grouping and extraction-program synthesis are inherently interdependent~\cite{crescenzi2001roadrunner,scribes2026,marineli2025controlplanes}. As illustrated in \Cref{fig:layout-rule-compatibility}, under anchor-based extraction~\cite{esser2014fewexemplar}, documents with similar layouts may require different rules due to variations in field labels (\cref{fig:layout-rule-compatibility}(a)), while documents with different layouts may share the same rule (\cref{fig:layout-rule-compatibility}(b)). These examples highlight that the suitability of a grouping depends on the extraction rules used to process its documents, while the extraction programs must, in turn, be synthesized from the documents assigned to each group.

Existing approaches often establish document groups based on predefined similarity measures, such as layout or semantic similarity, or domain-specific heuristics, such as vendor identities~\cite{ng2026aibaam,mittal2026billspp}. In pipelines that subsequently develop routing and extraction components under fixed group boundaries, document grouping is treated as a prerequisite for downstream development. Such a sequential design adapts the components to the grouping without revisiting whether the grouping is appropriate for their capabilities. Consequently, group boundaries that hinder reliable routing or program sharing may persist despite downstream optimization, ultimately limiting extraction accuracy.

To address these challenges, we propose EvoKIE, a co-evolutionary KIE framework that jointly refines document grouping, the router, and extraction programs. EvoKIE uses routing and extraction feedback to diagnose grouping issues and revise group boundaries. 
Specifically, routing errors guide document reassignment or group merging, while extraction failures motivate group splitting. To accommodate evolving group boundaries, we introduce an evaluation-driven agent optimization mechanism that optimizes the router and synthesizes group-specific extraction programs under the current grouping. At each iteration, the resulting KIE system, comprising the router and extraction programs developed under the refined grouping, is evaluated on a development set to guide subsequent regrouping, router optimization, and program synthesis.

We evaluate our framework on two public benchmarks~\cite{limam2025fatura,simsa2023docile} and on LogiInvoice, a proprietary dataset of multilingual logistics invoices. Our method achieves the highest field-level F1 score and document-level accuracy on the datasets. The largest improvement is observed on FATURA, where document-level accuracy increases from 65.00\% for the strongest baseline to 92.67\%, a gain of 27.67 percentage points.

Our main contributions are summarized as follows:
\begin{itemize}
    \item \textbf{Co-evolutionary KIE framework.} We propose EvoKIE, which iteratively refines document grouping, routing, and extraction programs. To the best of our knowledge, this is the first KIE framework to co-evolve all three components, enabling group boundaries and downstream routing and extraction strategies to adapt to one another.

    \item \textbf{Evaluation-driven module optimization.} To support this co-evolution, we introduce an agentic optimization mechanism that automatically optimizes the router and synthesizes group-specific extraction programs in coordination with evolving document groups.

    \item \textbf{Improved end-to-end extraction performance.} Experiments on three diverse document datasets demonstrate consistent improvements in field-level F1 and document-level accuracy over competitive baselines, with document-level accuracy gains of up to 27.67 percentage points.
\end{itemize}

\section{Related Work}
\label{sec:related-work}

\paragraph{Document Key Information Extraction.}
Document KIE has been extensively studied through supervised learning and, more recently, training-free approaches. Supervised methods integrate textual, visual, and spatial information, as exemplified by the LayoutLM family~\cite{xu2020layoutlm,xu2021layoutlmv2,huang2022layoutlmv3}, while Donut~\cite{kim2022donut} enables OCR-free document understanding. Recent approaches leverage pretrained language models without task-specific fine-tuning. ICL-D3IE~\cite{he2023icld3ie} iteratively refines in-context demonstrations, while SAIL~\cite{zhang2025sail} selects document-specific demonstrations based on textual and layout similarity. While these approaches adapt extraction through model training or in-context demonstrations, EvoKIE constructs reusable extraction programs and jointly refines document grouping to determine where they apply.

\paragraph{Program Reuse for Document Extraction.}
A complementary line of research exploits shared document structures to reuse extraction programs. LRSyn~\cite{parthasarathy2022landmarks} clusters documents using global blueprints and merges clusters based on local regions and landmarks to synthesize shared programs. TWIX~\cite{lin2026twix} infers visual templates from documents sharing an underlying format, while CACHE-ED2~\cite{bhoi2026cacheed2} uses a learned format encoder to associate documents with cached programs. These methods provide different ways to determine the scope of program reuse. EvoKIE explicitly addresses the interdependence between document grouping and module capabilities: routing and extraction feedback guide group revision, while the router and extraction programs adapt to the revised groups.

\paragraph{Evaluation-Driven Agent Optimization.}
Evaluation-driven agents improve candidate solutions through automated evaluation and feedback-guided revision. This paradigm has been explored in prompt optimization~\cite{agrawal2026gepa}, algorithm discovery~\cite{assumpcao2025codeevolve}, and agent skill refinement~\cite{yang2026skillopt}. In EvoKIE, regrouping changes both the routing targets and the extraction requirements of each program. We use evaluation-driven agent optimization to automatically adapt the router and extraction programs to these changes, enabling their co-evolution with document grouping.

\section{Method}
\label{sec:method}

\subsection{Framework Overview}
\label{sec:method-overview}
\begin{figure*}[t]
    \centering
    \includegraphics[width=\textwidth,trim=15pt 27pt 7pt 33pt,clip]
        {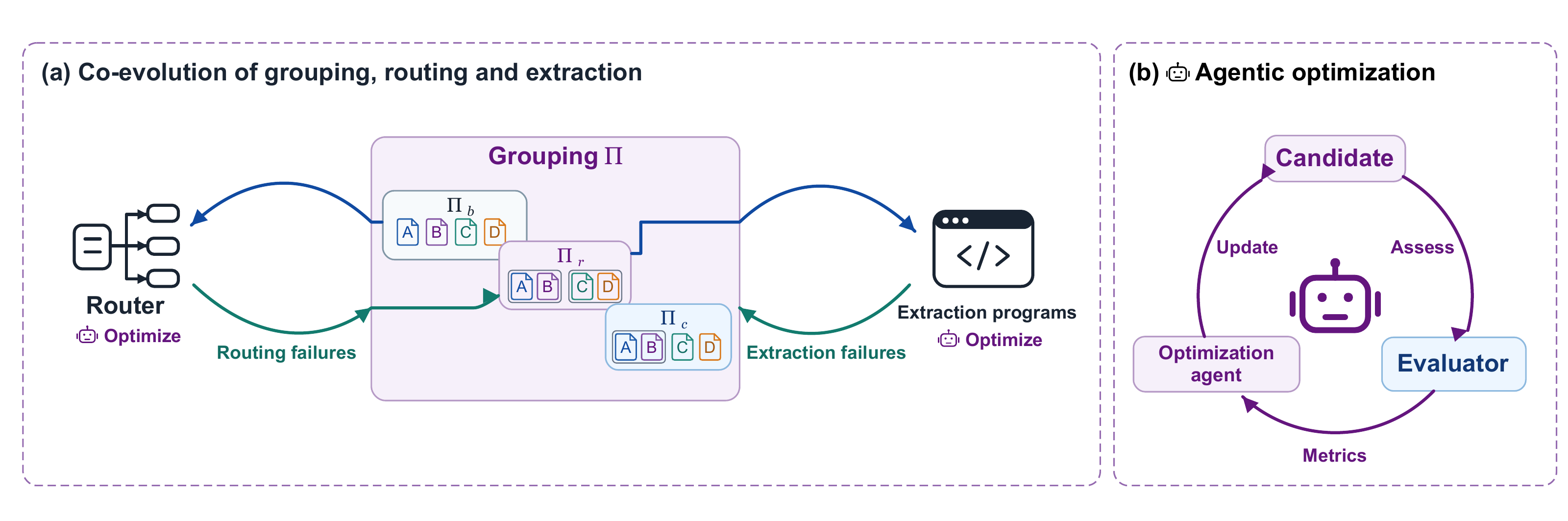}
    \caption{\textbf{Overview of EvoKIE.}
    (a) Co-evolution of grouping, routing and extraction. Grouping shapes routing and extraction-program synthesis,
while feedback from both modules drives iterative regrouping.
    (b) Agentic module optimization mechanism for the router and
    extraction program.}
    \label{fig:evokie-framework}
\end{figure*}

\paragraph{Problem setup.}
Given a document $x$ and a predefined field set $\mathcal F$,
key information extraction (KIE) predicts the value
$\hat y_f(x)$ of each field $f\in\mathcal F$,
with ground truth $y_f(x)$.

\paragraph{Co-evolutionary framework.}
Our framework consists of a document grouping $\Pi$,
a router $R$, and group-specific extraction programs
$\{p_g\}_{g\in\Pi}$(Fig.~\ref{fig:evokie-framework}).
Each program extracts predefined fields from applicable
documents and may reject inputs outside its scope~\cite{geifman2017selective}.

We maintain the evolving system as a history $\mathcal H$
of versions, each defined as
\[
V=\bigl(\Pi,R,\{p_g\}_{g\in\Pi}\bigr).
\]
Starting from an initial partition $\Pi_0$ of the labeled
training set $\mathcal D_{\mathrm{tr}}$, we construct
$V_0$, evaluate it on the development set
$\mathcal D_{\mathrm{dev}}$, and initialize
$\mathcal H=\{V_0\}$.

Each subsequent round selects a parent version from
$\mathcal H$ and alternates between module optimization
and feedback-driven regrouping.
Specifically, each round sequentially performs
router optimization, routing-driven regrouping,
program synthesis, and extraction-driven regrouping.
Through this process, feedback from routing and extraction
guides grouping adaptation
(Sec.~\ref{sec:method-regrouping}),
while the evolving partition shapes subsequent
router and program optimization
(Sec.~\ref{sec:method-agent}).
The resulting version is evaluated on
$\mathcal D_{\mathrm{dev}}$ and added to $\mathcal H$.
At inference time, the router directs each document
to a group-specific extraction program.

\paragraph{System-level evaluation and version selection.}
We evaluate each system version on the development set
$\mathcal D_{\mathrm{dev}}$.
Since extraction programs may reject documents outside
their applicable scope, let
$\mathcal C(V)\subseteq\mathcal D_{\mathrm{dev}}$
denote the set of documents with accepted extraction outputs.
We measure macro-averaged field F1 $F(V)$ and
document-level accuracy $A(V)$ on $\mathcal C(V)$,
and define program coverage as
$C(V)=|\mathcal C(V)|/|\mathcal D_{\mathrm{dev}}|$.

We define a hierarchical version selection score
$S(V)$ based on the three metrics for the 
final version selection:
\[
S(V)=
\begin{cases}
\bigl(1,C(V)\bigr),
& F(V)\ge\tau_F,\ A(V)\ge\tau_A,\\
\bigl(0,\frac{F(V)+A(V)}{2}\bigr),
& \text{otherwise}.
\end{cases}
\]
Scores are compared lexicographically, prioritizing
extraction quality thresholds before maximizing coverage.

\paragraph{Parent selection for co-evolution.}
We adopt a greedy single-parent search strategy for co-evolution. At the beginning of each round, we select the historical version with the highest $S(V)$ as the parent and provide its full development-set metrics to the agent. Inspired by AlphaEvolve~\cite{novikov2025alphaevolve}, we additionally provide the top three historical versions ranked by $S(V)$, along with their development-set metrics, as read-only references to encourage diverse revisions.

\paragraph{Compatibility and generalization diagnostics.}
To identify potential incompatibility and overfitting
during co-evolution, we provide the agent with
complementary routing and extraction diagnostics
derived from training and development performance.

For routing--extraction compatibility, we examine
the held-out Top-1 routing accuracy $T_r(R;\Pi)$
obtained during router optimization and
extraction-program acceptance on the development set.
The former measures how well the router distinguishes
the group labels defined by the current partition $\Pi$.

For the latter, we measure whether routed documents
are accepted by their assigned extraction programs.
Let $\mathcal R_g(V)$ denote the development-set
documents routed to group $g$ by version $V$.
The routing-conditioned program acceptance rate is
\[
\rho_g^{\mathrm{dev}}(V)=
\frac{|\mathcal R_g(V)\cap\mathcal C(V)|}
{|\mathcal R_g(V)|},
\]
for nonempty $\mathcal R_g(V)$.
High routing accuracy accompanied by low program
acceptance may indicate incompatibility between
routing decisions and extraction-program applicability.

For extraction generalization, we evaluate each
group-specific program $p_g$ on accepted in-group
training documents and development documents
whose outputs are produced by $p_g$.
We denote the corresponding field F1 and
document-level accuracy by
$(F_g^{\mathrm{tr}},A_g^{\mathrm{tr}})$ and
$(F_g^{\mathrm{dev}},A_g^{\mathrm{dev}})$,
respectively. When both sets contain evaluable
documents, we compute the train--development gaps:
\[
\Delta_g^F=F_g^{\mathrm{tr}}-F_g^{\mathrm{dev}},
\qquad
\Delta_g^A=A_g^{\mathrm{tr}}-A_g^{\mathrm{dev}}.
\]
Large positive gaps may indicate overfitting to
document-specific extraction rules that fail to
generalize to unseen documents.

\begin{algorithm}[t]
\caption{EvoKIE System}
\label{alg:kie-framework}
\small
\begin{algorithmic}[1]
\Require $\mathcal D_{\mathrm{tr}},\mathcal D_{\mathrm{dev}},
\Pi_0,\tau_F,\tau_A$
\Ensure System version $V^\star$

\Statex \textbf{Initialization}
\State $R_0\gets
\operatorname{Optimize}(R_{\mathrm{init}};\Pi_0,\mathcal E_r)$
\For{each group $g\in\Pi_0$}
    \State $p_{0,g}\gets
    \operatorname{Optimize}(p_{g,0};\Pi_0,\mathcal E_{c,g})$
\EndFor
\State $V_0\gets(\Pi_0,R_0,\{p_{0,g}\}_{g\in\Pi_0})$
\State Evaluate $V_0$ on $\mathcal D_{\mathrm{dev}}$
\State $\mathcal H\gets\{V_0\}$

\Statex \textbf{Co-evolution}
\While{stopping condition is unmet}
    \State Select parent
    $V_b=(\Pi_b,R_b,\{p_{b,g}\}_{g\in\Pi_b})$ from $\mathcal H$

    \State $\widetilde R_b\gets
    \operatorname{Optimize}(R_b;\Pi_b,\mathcal E_r)$
    \State Obtain routing feedback
    $E_r=(T_r,\mathcal B_r)$
    \State $\Pi_r\gets
    \mathcal G(\Pi_b,E_r,
    \{\mathrm{move},\mathrm{merge}\})$

    \For{each group $g\in\Pi_r$}
        \State $p_g\gets
        \operatorname{Optimize}(p_{g,0};\Pi_r,\mathcal E_{c,g})$
        \State Obtain execution feedback
        $E_{c,g}=(F_g,A_g,\mathcal B_{c,g})$
    \EndFor

    \State $\Pi_c\gets
    \mathcal G(\Pi_r,\{E_{c,g}\}_{g\in\Pi_r},
    \{\mathrm{split}\})$

    \State Rebuild routing index under $\Pi_c$
    \For{each group $g\in\Pi_c$}
        \State $p_{c,g}\gets
        \operatorname{Optimize}(p_{g,0};\Pi_c,\mathcal E_{c,g})$
    \EndFor

    \State $V_c\gets
    (\Pi_c,R_c,\{p_{c,g}\}_{g\in\Pi_c})$
    \State Evaluate $V_c$ on $\mathcal D_{\mathrm{dev}}$
    \State $\mathcal H\gets\mathcal H\cup\{V_c\}$
\EndWhile

\State Select $V^\star$ by coverage among
versions satisfying quality thresholds
\end{algorithmic}
\end{algorithm}

\subsection{Routing- and Extraction-Guided Regrouping}
\label{sec:method-regrouping}

To adapt document grouping to updated routing and
extraction strategies, we employ a regrouping agent
that refines document partitions based on routing
and program execution feedback. Let \(\mathcal G(\Pi,E,\mathcal O)\) denote a partition update based on feedback \(E\) and allowed operations \(\mathcal O\). Each round starts from a selected parent version \(V_b=(\Pi_b,R_b,\{p_{b,g}\}_{g\in\Pi_b})\).

\paragraph{Routing-driven adaptation.}
We optimize the parent router \(R_b\) under \(\Pi_b\) (Sec.~\ref{sec:method-agent}) to obtain \(\widetilde R_b\). Its Top-1 accuracy \(T_r\) and routing errors \(\mathcal B_r\) form the feedback \(E_r=(T_r,\mathcal B_r)\). The regrouping agent uses routing confusion to guide document reassignment or group merging, subject to extraction-rule compatibility:
\[
\Pi_r=
\mathcal G\!\left(
\Pi_b,E_r,\{\mathrm{move},\mathrm{merge}\}
\right).
\]

\paragraph{Extraction-driven adaptation.}
Under \(\Pi_r\), we optimize a shared extraction program \(p_g\) for each group \(g\in\Pi_r\) (Sec.~\ref{sec:method-agent}). Macro-averaged field F1 \(F_g\), document-level accuracy \(A_g\), and extraction errors \(\mathcal B_{c,g}\) form the feedback \(E_{c,g}=(F_g,A_g,\mathcal B_{c,g})\). The regrouping agent splits groups whose execution failures indicate incompatible extraction rules:
\[
\Pi_c=
\mathcal G\!\left(
\Pi_r,\{E_{c,g}\}_{g\in\Pi_r},
\{\mathrm{split}\}
\right).
\]

Given the refined partition \(\Pi_c\), we rebuild the routing index using the fixed routing procedure and optimize group-specific extraction programs, yielding the updated system version \(V_c=(\Pi_c,R_c,\{p_{c,g}\}_{g\in\Pi_c})\).

\subsection{Agentic Optimization of Routing and Extraction Programs}
\label{sec:method-agent}

As document partitions evolve, the router must identify document representations and routing strategies that effectively discriminate among groups defined by the updated labels, while program synthesis
must infer shared extraction rules from the documents
assigned to each refined group. To automate this adaptation, we employ a unified agentic module optimization mechanism for both modules.

\paragraph{Agentic module optimization.}
Given a fixed partition $\Pi$, we formulate module
optimization as an agent-guided search over candidate
solutions, with an optimization agent proposing
revisions and an independent evaluator $\mathcal E$ 
providing performance feedback.
At iteration $t$, the evaluator $\mathcal E$ assesses
candidate $z_t$ to produce feedback $e_t$, which,
together with accumulated experimental insights
maintained in a notebook~\cite{shinn2023reflexion,zhang2026ace} $\mathcal M_t$, informs
the agent's subsequent candidate revision:
\[
e_t=\mathcal E(z_t),\qquad
z_{t+1}=\operatorname{Agent}(z_t,e_t,\mathcal M_t).
\]
We denote the selected candidate by
$\operatorname{Optimize}(z_0;\Pi,\mathcal E)$,
where $z_0$ is the initial candidate.

\paragraph{Router optimization.}
Given a partition $\Pi$, router optimization searches
for document representations, index construction
methods, and retrieval strategies that effectively
discriminate among the assigned group labels.
To expand the agent's search space, we provide three
document representation primitives---OCR-based text
extraction~\cite{cui2026ppocrv5}, BGE-based text
encoding~\cite{xiao2024cpack}, and SigLIP-based
visual encoding~\cite{zhai2023siglip,kanagalingam2026invoicehaystack}---which
can be composed into candidate routing strategies.

To reduce sensitivity to individual data splits,
the evaluator $\mathcal E_r$ assesses each router
candidate $R_t$ over $K=10$ independent random
train-validation splits~\cite{arlot2010crossvalidation}.
For each split $k$, the training set is partitioned into
an index construction set $I_k$ and a held-out query
set $Q_k$, such that
\[
I_k\cap Q_k=\varnothing,\qquad
I_k\cup Q_k=\mathcal D_{\mathrm{tr}}.
\]
The routing index is constructed exclusively from
$I_k$, while documents in $Q_k$ serve as unseen queries.
For each query $x\in Q_k$, the router predicts the
top-ranked group $\hat g_{R_t,k}(x)$, which is compared
against the group label $g_\Pi(x)$ assigned by the
current partition $\Pi$.
The held-out Top-1 routing accuracy is computed as
\[
\operatorname{Acc}_{\mathrm{Top1}}^{(k)}(R_t;\Pi)
=
\frac{1}{|Q_k|}
\sum_{x\in Q_k}
\mathbf 1[
\hat g_{R_t,k}(x)=g_\Pi(x)
].
\]
The evaluator aggregates the accuracy across all
splits with equal weight to obtain
\[
\mathcal E_r(R_t)
=
T_r(R_t;\Pi)
=
\frac{1}{K}\sum_{k=1}^{K}
\operatorname{Acc}_{\mathrm{Top1}}^{(k)}(R_t;\Pi).
\]
This repeated held-out evaluation reduces dependence
on a particular index construction and query split,
providing a more stable assessment of the router's
ability to distinguish groups under the current
partition.

\paragraph{Extraction program synthesis.}
Given a partition $\Pi$, extraction program synthesis
aims to capture shared extraction rules within each
group $g\in\Pi$, enabling accurate in-group extraction
and out-of-group rejection. This allows programs to identify documents beyond their applicable scope rather than produce potentially incorrect outputs, enabling fallback to VLM-based extraction or manual review in practical enterprise applications.
It also provides a safeguard against
routing errors by rejecting misrouted documents.

For each group $g$, the optimization set consists of
labeled in-group examples $\mathcal D_g^{\mathrm{in}}$
and confusable out-of-group examples
$\mathcal D_g^{\mathrm{out}}$.
The in-group set contains the training documents
assigned to $g$ under the current partition:
\[
\mathcal D_g^{\mathrm{in}}
=
\{(x,y)\in\mathcal D_{\mathrm{tr}} \mid \Pi(x)=g\}.
\]

To construct the out-of-group set, we use in-group
documents as queries to retrieve similar documents
from other groups through the router optimized in
the preceding step of the current iteration.
These hard negatives reflect the confusions of the
current router, ensuring that program synthesis
is aligned with its routing behavior and encouraging
the program to reject documents for which its
extraction rules are not applicable.

The evaluator $\mathcal E_{c,g}$ measures field F1
and document-level accuracy on accepted in-group
documents, together with in-group coverage
$C_g(p_{g,t})$ and out-of-group false acceptance
rate $\operatorname{FAR}_g(p_{g,t})$:
\[
\begin{aligned}
C_g(p_{g,t})
&=
\frac{1}{|\mathcal D_g^{\mathrm{in}}|}
\sum_{x\in\mathcal D_g^{\mathrm{in}}} a_{g,t}(x),\\
\operatorname{FAR}_g(p_{g,t})
&=
\frac{1}{|\mathcal D_g^{\mathrm{out}}|}
\sum_{x\in\mathcal D_g^{\mathrm{out}}} a_{g,t}(x),
\end{aligned}
\]
where $a_{g,t}(x)=1$ if candidate $p_{g,t}$
accepts document $x$, and $0$ otherwise.
Together, these metrics encourage programs to
maintain high extraction accuracy and in-group
coverage while minimizing false acceptance
of out-of-group documents.

\section{Experiments}
\label{sec:experiments}

\subsection{Experimental Setup}
\label{sec:experimental_setup}

\paragraph{Datasets.}
We evaluate our method on two public benchmarks, FATURA and DocILE, and a private industrial dataset, LogiInvoice.
\textbf{FATURA}~\cite{limam2025fatura} comprises invoices spanning 50 distinct layouts, with 31 target fields, such as invoice number, seller name, and invoice total.
\textbf{DocILE}~\cite{simsa2023docile} consists of diverse business documents, including invoices, receipts, purchase orders, and utility bills, with a long-tailed layout distribution and 36 target fields, such as vendor name, issue date, and gross total amount.
\textbf{LogiInvoice} comprises multilingual logistics invoices collected from real-world industrial applications across multiple countries, with 14 target fields, including tax, currency, vendor address, and due date.

\paragraph{Data Splits.}
We split each dataset into training, development, and test sets at the ratio of 2:2:1. The training and development sets are used for system construction and optimization, while the test set is held out for final evaluation.

\paragraph{Evaluation Metrics.}
We report macro-averaged field accuracy (FA), field F1 (F1), and document-level accuracy (DA).
Let $N$ denote the number of test documents and $\mathcal F$ the set of target fields.
For each field $f$, let $m_f$ denote the number of correctly matched nonempty predictions, $z_f$ the number of prediction--reference pairs that are both empty, and $p_f$ and $g_f$ the numbers of nonempty predictions and references, respectively.
The metrics are computed as
\begin{align}
\mathrm{FA} &= \frac{1}{|\mathcal F|}\sum_{f\in\mathcal F}\frac{m_f+z_f}{N},\\
\mathrm{F1} &= \frac{1}{|\mathcal F|}\sum_{f\in\mathcal F}\frac{2m_f}{p_f+g_f},\\
\mathrm{DA} &= \frac{1}{N}\sum_{i=1}^{N}
\mathbb{I}\left[
\forall f\in\mathcal F,\,
\hat y_f(x_i)=y_f(x_i)
\right].
\end{align}

\paragraph{Implementation Details.}
The agents for regrouping, router optimization, and extraction program synthesis use the Pi Agent framework~\cite{piagent} with Qwen3.8-Max.

\subsection{Main Results}
\begin{table*}[t]
    \centering
    \caption{Test-set extraction performance (\%). Each dataset reports macro field accuracy (FA), macro field F1 (F1), and document-level accuracy (DA). Bold indicates the highest value in each column.}
    \label{tab:main_results}
    \begin{tabular*}{\textwidth}{@{\extracolsep{\fill}}l*{9}{r}@{}}
        \toprule
        & \multicolumn{3}{c}{FATURA}
        & \multicolumn{3}{c}{DocILE}
        & \multicolumn{3}{c}{LogiInvoice} \\
        \cmidrule(lr){2-4}
        \cmidrule(lr){5-7}
        \cmidrule(l){8-10}
        Method & FA & F1 & DA & FA & F1 & DA & FA & F1 & DA \\
        \midrule
        \rowcolor[gray]{0.92}
        \multicolumn{10}{l}{\textit{Zero-shot VLM}} \\
        Qwen3.8-Max
        & 95.31 & 83.07 & 24.67
        & 86.75 & 42.11 & 0.83
        & 96.39 & 96.68 & 59.60 \\
        Kimi-K3.1-Pro
        & 97.40 & 89.40 & 57.00
        & 87.73 & 47.96 & 2.21
        & 97.69 & 97.90 & 74.75 \\
        GPT-6.1-Sol
        & 97.73 & 91.42 & 65.00
        & 84.77 & 44.76 & 1.94
        & 97.62 & 98.09 & 74.75 \\
        \midrule
        \rowcolor[gray]{0.92}
        \multicolumn{10}{l}{\textit{In-context Learning}} \\
        SAIL
        & 97.08 & 91.10 & 29.33
        & 92.74 & 75.37 & 17.96
        & 83.19 & 89.01 & 73.74 \\
        \midrule
        \rowcolor[gray]{0.92}
        \multicolumn{10}{l}{\textit{Program-based}} \\
        LRSyn
        & 87.80 & 78.88 & 0.67
        & 74.88 & 48.18 & 0.00
        & 66.67 & 69.45 & 12.12 \\
        TWIX
        & 63.31 & 25.81 & 0.00
        & 75.00 & 35.57 & 0.00
        & 16.27 & 11.83 & 0.00 \\
        Ours
        & \textbf{99.67} & \textbf{98.88} & \textbf{92.67}
        & \textbf{93.49} & \textbf{78.49} & \textbf{25.14}
        & \textbf{99.06} & \textbf{99.08} & \textbf{89.90} \\
        \bottomrule
    \end{tabular*}
\end{table*}

\paragraph{Baselines.}
We compare against direct extraction with three general-purpose MLLMs (Qwen~\cite{alibaba2026qwen38max}, Kimi~\cite{moonshot2026kimik31pro}, and GPT~\cite{openai2026gpt61sol}), in-context extraction with SAIL~\cite{zhang2025sail}, and program-based extraction with LRSyn~\cite{parthasarathy2022landmarks} and TWIX~\cite{lin2026twix}. SAIL selects document-specific demonstrations by textual and layout similarity. LRSyn groups documents by structural similarity to synthesize shared extraction programs, while TWIX infers visual templates from documents assigned to predefined groups. We also consider CACHE-ED2~\cite{bhoi2026cacheed2}, which requires a separately trained neural format classifier for document routing. However, the classifier's pretrained weights are not publicly available, making faithful reproduction difficult. Although CACHE-ED2 also evaluates on FATURA, it does not report comparable end-to-end KIE accuracy, precluding a direct comparison using its published results.

\paragraph{Results.}
Our method achieves the highest scores on all three metrics across all three datasets (Table~\ref{tab:main_results}). It improves document-level accuracy over the strongest reported baseline by 27.67 percentage points on FATURA, 7.18 on DocILE, and 15.15 on LogiInvoice. DocILE particularly illustrates the gap between field-level and document-level performance: the best zero-shot MLLM
achieves 87.73\% field accuracy, yet only 2.21\% of its
documents have all fields correctly extracted. Since document-level accuracy requires all fields
to be correct, it tends to decay exponentially
as the number of target fields increases,
making this challenge particularly pronounced
on DocILE with 36 target fields. Our method raises these scores to 93.49\% and 25.14\%, respectively.

\subsection{Ablation Study and Analysis}

\begin{figure}[t]
    \centering
    \includegraphics[width=\columnwidth]{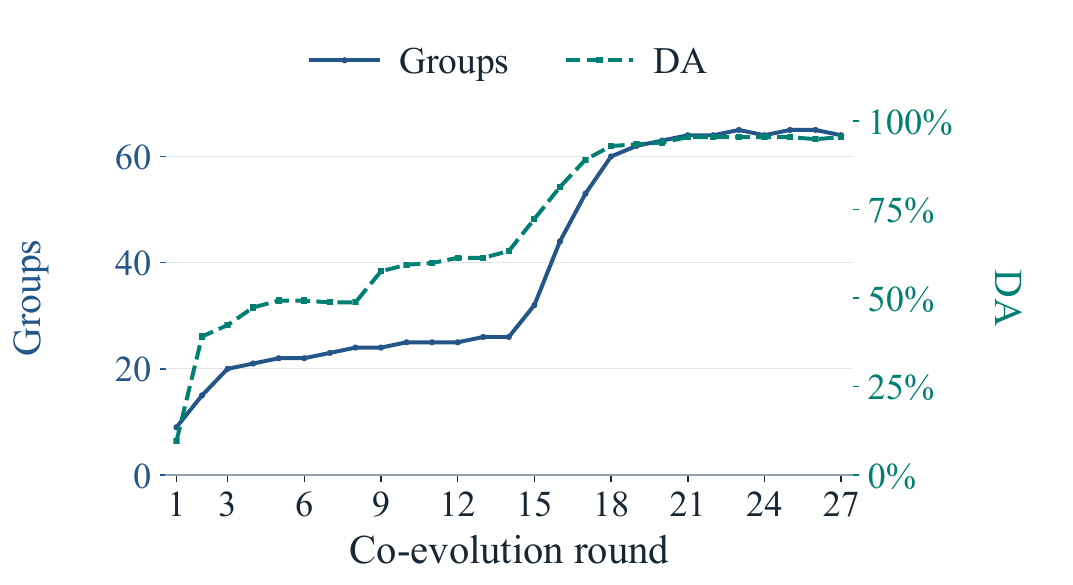}
    \caption{Number of document groups (left axis) and development-set
    document-level accuracy (right axis) over 27 co-evolution rounds
    on LogiInvoice.}
    \label{fig:group_evolution}
\end{figure}

\begin{table}[t]
\centering
\caption{Effect of adaptive document grouping on LogiInvoice. FA, F1, and DA follow Table~\ref{tab:main_results}. Cov.\ and Apt.\ DA denote program acceptance coverage and accepted-document accuracy, respectively.}
\label{tab:group_ablation}
\small
\setlength{\tabcolsep}{4pt}
\begin{tabular}{@{}ccrrrrr@{}}
\toprule
Routing & Extraction & FA & F1 & DA & Cov. & Apt.\ DA \\
\midrule
\(\times\) & \(\times\) & 96.10 & 96.74 & 69.70 & 17.17 & 58.82 \\
\(\checkmark\) & \(\times\) & 88.96 & 90.08 & 39.39 & 100.00 & 39.39 \\
\(\times\) & \(\checkmark\) & 98.02 & 98.88 & 87.88 & 91.92 & 89.01 \\
\(\checkmark\) & \(\checkmark\) & \textbf{99.06} & \textbf{99.08} & \textbf{89.90} & 92.93 & \textbf{89.13} \\
\bottomrule
\end{tabular}
\end{table}

\paragraph{Effect of Adaptive Document Grouping.}
We evaluate adaptive document grouping on LogiInvoice by independently enabling regrouping driven by routing and extraction feedback. Since EvoKIE iteratively refines document grouping, ground-truth group assignments for test documents cannot be determined in advance, making it difficult to evaluate the routing and extraction modules separately. We therefore use two proxy metrics to approximate their respective performance: program acceptance coverage (Coverage) for routing and accepted-document accuracy (AcceptedDA) for extraction. For a system version $V$ evaluated on the test set $\mathcal D_{\mathrm{test}}$, let $\mathcal C_{\mathrm{test}}(V)$ denote the set of documents with accepted extraction outputs, following the definition in Sec.~\ref{sec:method}. The two metrics are defined as

\begin{align}
\mathrm{Coverage} &=
\frac{|\mathcal C_{\mathrm{test}}(V)|}
{|\mathcal D_{\mathrm{test}}|},\\
\mathrm{AcceptedDA} &=
\frac{
\sum_{d\in\mathcal C_{\mathrm{test}}(V)}
\mathbb I[\forall f\in\mathcal F,\hat y_f(d)=y_f(d)]
}{
|\mathcal C_{\mathrm{test}}(V)|
}.
\end{align}

As shown in Table~\ref{tab:group_ablation}, jointly enabling both feedback mechanisms improves DA from 69.70\% with static grouping to 89.90\%, outperforming either mechanism alone. The two proxy metrics reveal contrasting effects: routing-driven regrouping alone achieves higher Coverage (100.00\% vs. 91.92\%) but substantially lower AcceptedDA (39.39\% vs. 89.01\%) than extraction-driven regrouping alone. This suggests that optimizing grouping for routing compatibility alone may compromise extraction quality, highlighting the competing requirements of the two modules. Notably, extraction-driven regrouping achieves a DA of 87.88\%, approaching the joint configuration (89.90\%), whereas routing-driven regrouping alone yields only 39.39\%. This asymmetry suggests that extraction compatibility has a greater impact on end-to-end performance, potentially because routing errors can be mitigated through program rejection and few-shot fallback, whereas extraction-rule conflicts may produce incorrect but accepted outputs that bypass fallback.
These results demonstrate that static grouping or refining grouping for either module in isolation is suboptimal, while jointly incorporating both feedback signals enables EvoKIE to achieve the best overall extraction performance.

\begin{table}[t]
\centering
\caption{Effect of agentic module optimization on LogiInvoice. Check marks indicate iterative agentic optimization, while crosses indicate single-pass agent generation without an explicit evaluation-feedback loop.}
\label{tab:agent_ablation}
\begin{tabular}{@{}ccrrrr@{}}
\toprule
Router & Program & FA & F1 & DA & Cov. \\
\midrule
\(\times\)    & \(\times\)    & 98.77 & 98.72 & 84.85 & 89.90 \\
\(\checkmark\) & \(\times\)    & 87.16 & 88.61 & 44.44 & 98.99 \\
\(\times\)    & \(\checkmark\) & 98.92 & 99.02 & \textbf{89.90} & 88.89 \\
\(\checkmark\) & \(\checkmark\) & \textbf{99.06} & \textbf{99.08} & \textbf{89.90} & 92.93 \\
\bottomrule
\end{tabular}
\end{table}

\paragraph{Effect of Agentic Module Optimization.}
We evaluate iterative agentic optimization by replacing each optimization loop with single-pass agent generation while retaining adaptive grouping. As shown in Table~\ref{tab:agent_ablation}, enabling both loops improves DA from 84.85\% to 89.90\% (+5.05 points) over the single-pass baseline, achieving the highest field-level scores.

Nevertheless, single-pass agent generation performs better than expected, substantially outperforming competitive baselines (Table~\ref{tab:main_results}). To understand this observation, we inspect its execution traces and observe that agents spontaneously evaluate and refine their generated solutions without explicit instructions, suggesting a capacity for self-directed optimization.

Notably, enabling only router agentic optimization reduces DA from 84.85\% to 44.44\%, despite increasing Coverage from 89.90\% to 98.99\%. Inspection of the co-evolution traces reveals that this configuration produces only 20 final groups, compared with approximately 60 in the other configurations. One group contains substantially more documents than the others, with a size $9.5$ times the mean group size. Moreover, this group accounts for 97.6\% of the additional extraction failures relative to the single-pass baseline. These observations suggest that router optimization may favor overly coarse grouping, while single-pass program synthesis fails to adequately accommodate the resulting extraction-rule conflicts, providing insufficient feedback for subsequent group splitting. 

This result highlights the importance of coordinated agentic module optimization: improving routing effectiveness via the agentic mechanism in isolation may compromise extraction quality and even degrade end-to-end performance below the single-pass baseline. Iterative agentic optimization jointly enables the router and extraction programs to co-adapt through evolving document grouping, achieving the best overall performance.

\begin{table}[t]
\centering
\caption{Sensitivity to initial grouping granularity on
LogiInvoice. All performance metrics are percentages.}
\label{tab:granularity}
\small
\begin{tabular*}{\columnwidth}{@{\extracolsep{\fill}}llrrrrr@{}}
\toprule
& & \multicolumn{2}{c}{\# Groups}
& \multicolumn{3}{c}{Performance (\%)} \\
\cmidrule(lr){3-4}\cmidrule(l){5-7}
Granularity & $\varepsilon$ & Initial & Final & FA & F1 & DA \\
\midrule
Fine   & P50 & 9 & 68 & 98.85 & 98.86 & 89.90 \\
Medium & P75 & 3 & 64 & 99.06 & 99.08 & 89.90 \\
Coarse & P90 & 2 & 67 & 98.54 & 98.68 & 89.10 \\
\bottomrule
\end{tabular*}
\end{table}

\begin{table}[t]
\centering
\caption{Pairwise adjusted Rand index (ARI) across grouping
configurations before and after co-evolution.
$\Delta$ denotes the change in ARI.}
\label{tab:granularity_ari}
\small
\begin{tabular*}{\columnwidth}{@{\extracolsep{\fill}}lrrr@{}}
\toprule
Configuration pair & Initial ARI & Final ARI & $\Delta$ \\
\midrule
Fine--Medium   & 0.092 & 0.824 & $+0.732$ \\
Fine--Coarse   & 0.034 & 0.782 & $+0.748$ \\
Medium--Coarse & 0.565 & 0.858 & $+0.293$ \\
\bottomrule
\end{tabular*}
\end{table}

\paragraph{Sensitivity to Initial Grouping Granularity.}
EvoKIE initializes document groups by applying
DBSCAN~\cite{ester1996dbscan} to document embeddings.
We vary the neighborhood radius $\varepsilon$ using
the 50th (P50), 75th (P75), and 90th (P90) percentiles
of the distance distribution to obtain fine, medium,
and coarse initial groupings with 9, 3, and 2 groups,
respectively.
We investigate the sensitivity to initialization
in terms of both extraction performance and the
resulting document partitions.

To measure partition similarity, we compute the
adjusted Rand index (ARI)~\cite{hubert1985comparing}
between each pair of configurations, using the same
training documents before and after co-evolution.

As shown in Table~\ref{tab:granularity},
despite substantial differences in initial group counts,
the three configurations evolve into similar final
group counts of 68, 64, and 67.
Table~\ref{tab:granularity_ari} further shows that
pairwise ARI increases consistently across all
configuration pairs, with final values ranging from
0.782 to 0.858.
Notably, the Fine--Coarse pair exhibits the largest
increase, from 0.034 to 0.782, indicating that
initially distinct partitions can evolve toward
substantially more consistent group assignments.

Furthermore, all three configurations achieve
comparable field-level and document-level extraction
performance, with document-level accuracy varying
by only 0.8 percentage points.
These results suggest that EvoKIE is
robust to initial grouping granularity, allowing it to evolve from simple initial settings through iterative feedback rather than relying on a carefully tuned initialization.

\paragraph{Analysis of Group Evolution.}
Figure~\ref{fig:group_evolution} shows the evolution of group count and development-set accuracy across co-evolution rounds. Both metrics improve overall, but their trajectories reveal different stages of co-evolution. In the early rounds, accuracy increases substantially despite relatively modest growth in group count, suggesting that improvements are not solely attributable to finer grouping but also involve module optimization. Around rounds 15--18, group count rises sharply from approximately 26 to 60, accompanied by a marked improvement in accuracy. This observation is consistent with the effectiveness of extraction-driven regrouping in Table~\ref{tab:group_ablation}, where extraction feedback helps identify and separate documents requiring incompatible extraction rules.

In later rounds, both metrics largely stabilize, with group count fluctuating around 60--66 and accuracy remaining near its peak. This behavior suggests that continued optimization can refine system performance without requiring further substantial increases in grouping granularity. Together, these trends illustrate the complementary roles of module optimization and adaptive grouping: the former improves routing and extraction under the current partition, while the latter revises group boundaries to better accommodate their evolving requirements.

\section{Conclusion}
\label{sec:conclusion}

In this paper, we proposed EvoKIE, a co-evolutionary framework for document key information extraction that jointly refines document grouping, routing, and extraction programs. By leveraging routing and extraction feedback to refine group boundaries and employing agentic optimization to adapt both modules to the updated grouping, EvoKIE enables these interdependent components to co-evolve. Experiments on three diverse document datasets demonstrate consistent improvements in field-level F1 and substantial gains in document-level accuracy, reaching up to 27.67 percentage points over the strongest baselines. Ablation studies further validate the effectiveness of adaptive grouping and coordinated agentic module optimization, highlighting their complementary roles in improving end-to-end KIE performance through co-evolution.

{\small
    \bibliographystyle{ieeenat_fullname}
    \bibliography{main}
}
\clearpage
\onecolumn
\appendix
\section{More Implementation Details}

\subsection{Initial Document Grouping}

\paragraph{Initial grouping.}
We construct the initial partition $\Pi_0$ of the training set $D_{\mathrm{tr}}$ using fused text and visual embeddings. For each document $x_i$, let $b_i$ and $s_i$ denote its BGE text embedding~\cite{xiao2024cpack} and SigLIP visual embedding~\cite{zhai2023siglip}, respectively. We form
\begin{equation}
    z_i = \frac{1}{\sqrt{2}}
    \left[
        \frac{b_i}{\|b_i\|_2}
        \,;\,
        \frac{s_i}{\|s_i\|_2}
    \right],
    \qquad X = \{z_i\}_{i=1}^{|D_{\mathrm{tr}}|}.
\end{equation}
We obtain the preliminary grouping by applying DBSCAN~\cite{ester1996dbscan} with cosine distance:
\begin{equation}
    \Pi = \operatorname{DBSCAN}
    \bigl(X,\varepsilon,\mathrm{MinPts}\bigr),
\end{equation}
We set $\mathrm{MinPts}=m$ and compute the neighborhood radius from the training embeddings:
\begin{equation}
    \varepsilon =
    Q_P\!\left(
        \left\{
            d_{\cos}\bigl(z_i,z_i^{(m)}\bigr)
        \right\}_{i=1}^{|D_{\mathrm{tr}}|}
    \right),
\end{equation}
where $z_i^{(m)}$ denotes the $m$-th nearest neighbor of $z_i$, with $z_i$ itself included in the neighbor ranking, and $Q_P$ denotes the $P$-th percentile. Finally, we assign each noise document to the group of its nearest non-noise document under cosine distance, yielding the initial partition $\Pi_0$.

\subsection{Few-shot MLLM Fallback}
We use a few-shot MLLM as a fallback extractor when the routed extraction program returns a rejection. At the end of each round, the demonstration bank is constructed from the current partition $\Pi_c$ by randomly sampling one training document and its ground-truth field values from each group. For a query document that triggers fallback, we retrieve the demonstration corresponding to its routed group and provide it to the MLLM together with the query document and the field extraction rules for extraction.

\subsection{Agentic Module Optimization Details}

\paragraph{Agentic module optimization.}
Router optimization and extraction program synthesis use the same agentic optimization procedure described in Sec.~3.3. The agent revises a candidate program and invokes \texttt{submit\_candidate}; the controller independently evaluates the submitted program and returns feedback. Table~\ref{tab:agentic-module} summarizes the configuration.
\begin{table}[!htbp]
\caption{Configuration for agentic module optimization.}
\label{tab:agentic-module}
\begingroup
\small
\setlength{\tabcolsep}{6pt}
\renewcommand{\arraystretch}{1.16}
\begin{tabularx}{\textwidth}{@{}>{\raggedright\arraybackslash}p{0.94in}>{\raggedright\arraybackslash}X>{\raggedright\arraybackslash}X@{}}
\toprule
\textbf{Item} & \textbf{Router Optimization} & \textbf{Extraction Program Synthesis} \\
\midrule
Task prompt & \multicolumn{2}{p{\dimexpr\textwidth-0.94in-12pt\relax}@{}}{See Appendix~\ref{sec:prompt-templates} for the shared task-prompt template
(\ref{sec:prompt-shared}) and module-specific instructions
(\ref{sec:prompt-router}, \ref{sec:prompt-extraction}).} \\
\midrule
Agent context &
\dataitem{Training documents in $D_{\mathrm{tr}}$.}
\dataitem{The partition $\Pi_b$ of the parent version $V_b=(\Pi_b,R_b,\{p_g\}_{g\in\Pi_b})$ selected from $\mathcal H$.}
\dataitem{Sub-training and held-out sub-validation splits of $D_{\mathrm{tr}}$.}
\dataitem{The candidate router program, initialized from $R_b$ and editable by the agent.}
\dataitem{Reference router programs from the top three versions in $\mathcal H$, ranked by the version-selection score.}
\dataitem{OCR, BGE, and SigLIP functions callable by the program.}
\dataitem{The experiment notebook, available for reading and writing.}
\dataitem{Metric feedback from local candidate evaluation and system-level evaluation on $D_{\mathrm{dev}}$.}
&
\dataitem{In-group documents $x\in D_g^{\mathrm{in}}$ and out-of-group documents $x\in D_g^{\mathrm{out}}$, for $g\in\Pi_r$.}
\dataitem{KIE ground truth $y_f(x)$ for $f\in\mathcal F$.}
\dataitem{The same field extraction rules as those specified in the MLLM baseline prompt.}
\dataitem{The candidate extraction program, initialized from a parent program $p_g$ ($g\in\Pi_b$) and editable by the agent.}
\dataitem{Reference extraction programs from the top three versions in $\mathcal H$, ranked by the version-selection score.}
\dataitem{Document-processing functions callable by the program, including OCR.}
\dataitem{The experiment notebook, available for reading and writing.}
\dataitem{Metric feedback from local candidate evaluation and system-level evaluation on $D_{\mathrm{dev}}$.}
\\
\midrule
Agent configuration & \multicolumn{2}{p{\dimexpr\textwidth-0.94in-12pt\relax}@{}}{\textbf{Tools:} \texttt{read}, \texttt{bash}, \texttt{edit}, \texttt{write}, \texttt{grep}, \texttt{find}, \texttt{ls}, and \texttt{submit\_candidate}. \textbf{Model:} Qwen3.8-Max. \textbf{Thinking mode:} off. The \texttt{submit\_candidate} tool invokes the controller to evaluate the current candidate program independently.} \\
\midrule
Experiment notebook & \multicolumn{2}{p{\dimexpr\textwidth-0.94in-12pt\relax}@{}}{The agent records the parent candidate, problem, supporting evidence, hypothesis, proposed change, and expected outcome before submission; afterward, it records observations, regressions, conclusions, and the next action.\par\smallskip The controller independently records candidate source, evaluation metrics, query/extraction results, and evidence paths.} \\
\midrule
Evaluation metrics & Held-out top-1 routing accuracy $T_r$, as defined in Sec.~3.3. & Field F1 and document-level accuracy on accepted in-group documents, in-group coverage $C_g$, and out-of-group false acceptance rate $\mathrm{FAR}_g$, as defined in Sec.~3.3. \\
\bottomrule
\end{tabularx}
\par\endgroup
\smallskip
\noindent\begin{minipage}{\textwidth}\footnotesize
\textit{Group correspondence.} After routing-guided regrouping, we initialize extraction programs by matching groups in $\Pi_r$ to those in the parent partition $\Pi_b$ using Hungarian matching.
\end{minipage}
\end{table}

\subsection{Regrouping Details}

\paragraph{Routing- and extraction-guided regrouping.}
Both routing-guided and extraction-guided regrouping use an agent to inspect training documents and evaluation evidence and propose partition updates. Routing-guided regrouping allows document reassignment among existing groups and group merging, whereas extraction-guided regrouping allows only within-group splitting.
Table~\ref{tab:agentic-regrouping} summarizes their configurations.

\begin{table}[!htbp]
\caption{Configuration for routing- and extraction-guided regrouping.}
\label{tab:agentic-regrouping}
\begingroup
\small
\setlength{\tabcolsep}{6pt}
\renewcommand{\arraystretch}{1.16}
\begin{tabularx}{\textwidth}{
@{}>{\raggedright\arraybackslash}p{0.94in}
>{\raggedright\arraybackslash}X
>{\raggedright\arraybackslash}X@{}}
\toprule
\textbf{Item} &
\textbf{Routing-guided} &
\textbf{Extraction-guided} \\
\midrule
Task prompt &
\multicolumn{2}{p{\dimexpr\textwidth-0.94in-12pt\relax}@{}}{
See Appendix~\ref{sec:routing-regouping} and ~\ref{sec:extraction-regouping} for the regrouping task prompts.} \\
\midrule
Agent context &
\dataitem{Training documents in $D_{\mathrm{tr}}$ and the parent partition $\Pi_b$.}
\dataitem{Sub-training and held-out sub-validation splits of $D_{\mathrm{tr}}$.}
\dataitem{The optimized router program $\widetilde R_b$.}
\dataitem{Evaluation metrics and failure cases for $\widetilde R_b$.}
&
\dataitem{In-group documents $x\in D_g^{\mathrm{in}}$ and out-of-group
documents $x\in D_g^{\mathrm{out}}$, for $g\in\Pi_r$.}
\dataitem{The extraction program $p_g$.}
\dataitem{Evaluation metrics and failure cases for $p_g$.}
\\
\midrule
Agent configuration &
\multicolumn{2}{p{\dimexpr\textwidth-0.94in-12pt\relax}@{}}{
\textbf{Tools:} \texttt{read}, \texttt{bash}, \texttt{edit},
\texttt{write}, \texttt{grep}, \texttt{find}, \texttt{ls}, and
\texttt{submit\_candidate}.
\textbf{Model:} Qwen3.8-Max.
\textbf{Thinking mode:} off.
The \texttt{submit\_candidate} tool invokes the controller to validate
the structured regrouping proposal independently.} \\
\bottomrule
\end{tabularx}
\par\endgroup
\end{table}

\clearpage
\subsection{Experiment Details}

\subsubsection{Baselines}
\label{sec:supp-baselines}

We adapt SAIL and TWIX from the authors' public code and
reimplement LRSyn for image documents. OCR results are obtained
using PaddleOCR v5~\cite{cui2026ppocrv5}. All model calls in SAIL and TWIX use
Qwen3.8-Max~\cite{alibaba2026qwen38max} with temperature 0 and thinking disabled.

\paragraph{SAIL.}
We construct the demonstration pool for
SAIL~\cite{zhang2025sail} from the training set. OCR blocks
are associated with target fields using native bounding-box
annotations on FATURA and DocILE, and value-to-OCR matching
on LogiInvoice. For each query document, we retrieve four
layout demonstrations and four document-level demonstrations,
together with four entity examples per query OCR block.
Layout similarity is computed from first-page box
representations resized to $512\times512$, while textual
similarity uses \texttt{all-MiniLM-L6-v2}\footnote{Model card: \url{https://huggingface.co/sentence-transformers/all-MiniLM-L6-v2}.} with its default
256-token input limit. The retrieved examples are assembled
into the extraction prompt; if the input exceeds 100k tokens,
we progressively reduce the number of demonstrations to fit
the input budget.

\paragraph{TWIX.}
We organize the training documents into layout groups before
applying TWIX~\cite{lin2026twix}. FATURA and DocILE use their
native layout labels. For LogiInvoice, documents are initially
grouped by vendor and then individually reviewed and manually
regrouped so that each group contains documents sharing the
same layout. We infer one template from all training documents
in each group and map its native keywords to the target schema.
This mapping uses the field specification and native extraction
outputs paired with ground truth from up to three training
documents per group, specifying whether each target field
retains the value alone or concatenates the key and value.
At test time, the reference layout label selects the frozen
template and mapping, and extraction proceeds without
additional model calls.

\paragraph{LRSyn.}
As the original implementation is not publicly available,
we reimplement the image-document pipeline of
LRSyn~\cite{parthasarathy2022landmarks}. Program selection
follows the original NDSyn algorithm, while string
transformation uses Microsoft's publicly available FlashFill
SDK\footnote{Microsoft PROSE SDK: \url{https://github.com/microsoft/prose}.}. To provide the required training field locations, we use
native annotations on FATURA and DocILE and value-to-OCR
matching on LogiInvoice. During synthesis, region paths are
limited to four movements, with beam size 24 and a base-path
limit of 20. We select the configuration on the development
set by evaluating five synthesis configurations under four
blueprint-distance thresholds, yielding 20 candidate settings.
Candidates are ranked by document accuracy, with ties broken
by nonempty-field micro-F1 and then field accuracy. This
procedure selects clustering thresholds of 0.12 for FATURA
and 0.36 for DocILE and LogiInvoice.

\subsubsection{Datasets}
\textbf{FATURA}~\cite{limam2025fatura}. We randomly sample 30 invoices from each of 50 layouts, yielding 1,500 documents.
\textbf{DocILE}~\cite{simsa2023docile}. Documents with multiple conflicting annotated values for the same target field are excluded, resulting in 1,808 documents spanning 52 layouts.
\textbf{LogiInvoice.} LogiInvoice comprises 498 multilingual logistics invoices from 50 vendors, spanning approximately 50 layouts with occasional within-vendor variations. All baselines are evaluated on the same training, development, and test splits.

\clearpage
\section{Failure Case Analysis of Existing Methods}
\label{sec:supp-results-analysis}
\begingroup
\newcommand{\panelheight}{1.30in}
\newcommand{\casepanel}[3]{%
\begin{minipage}[t]{.237\linewidth}\centering
\begin{minipage}[t][.34in][t]{\linewidth}\centering\fontsize{7.5}{8.3}\selectfont #1\par\end{minipage}\par
\includegraphics[width=\linewidth,height=\panelheight,keepaspectratio]{#2}\par
{\fontsize{7}{8}\selectfont #3\par}
\end{minipage}}
\newcommand{\cardtitle}[1]{\par\noindent\rule{\linewidth}{.5pt}\par\vspace{1pt}\noindent\colorbox{gray!12}{\parbox{\dimexpr\linewidth-2\fboxsep\relax}{\fontsize{8.5}{10}\selectfont #1}}\par\vspace{1pt}}
\newcommand{\cardtext}[2]{\par\vspace{3pt}{\fontsize{8.5}{9.8}\selectfont\noindent\textbf{#1}\ #2\par}}
\definecolor{EvidenceGreen}{HTML}{087743}
\newcommand{\legendbox}[2]{\begingroup\setlength{\fboxsep}{0pt}\setlength{\fboxrule}{.8pt}\textcolor{#1}{\fbox{\rule{0pt}{5pt}\rule{10pt}{0pt}}}\endgroup\, #2}
\newcommand{\legenddivider}{\par\vspace{1pt}\noindent\rule{\linewidth}{.4pt}\par}
\captionsetup[figure]{font=casecaption,skip=8pt,justification=justified,singlelinecheck=false}
\setlength{\intextsep}{1pt}
\subsection{In-context Learning Method}
SAIL~\cite{zhang2025sail} selects in-context demonstrations based on textual and layout similarity and uses them to guide field extraction. However, similarity does not necessarily imply extraction-rule compatibility. Figure~\ref{fig:grouped-sail} illustrates two limitations: SAIL may struggle to consistently apply the appropriate extraction rule to the target document when the retrieved examples follow different rules (a,b), or fail to retrieve available references that follow the extraction rule required by the target (c). Through agentic module optimization, EvoKIE synthesizes a shared extraction program that accommodates different sub-layouts and their extraction rules within each group. Router optimization directs incoming documents to the corresponding group-specific program, whose extraction logic covers the relevant conditions and label variants rather than relying on the rules represented by a few retrieved examples. This reduces the risk of missing applicable rules during example retrieval.
\begin{figure}[H]
\renewcommand{\panelheight}{1.08in}
\cardtitle{\textbf{(a)}\quad DocILE\_697395\quad Ground truth: \texttt{\{"customer\_other\_name": "DICK GARRISON"\}}}
\casepanel{\textbf{SAIL target}\\ \textbf{Prediction:} \texttt{null}}{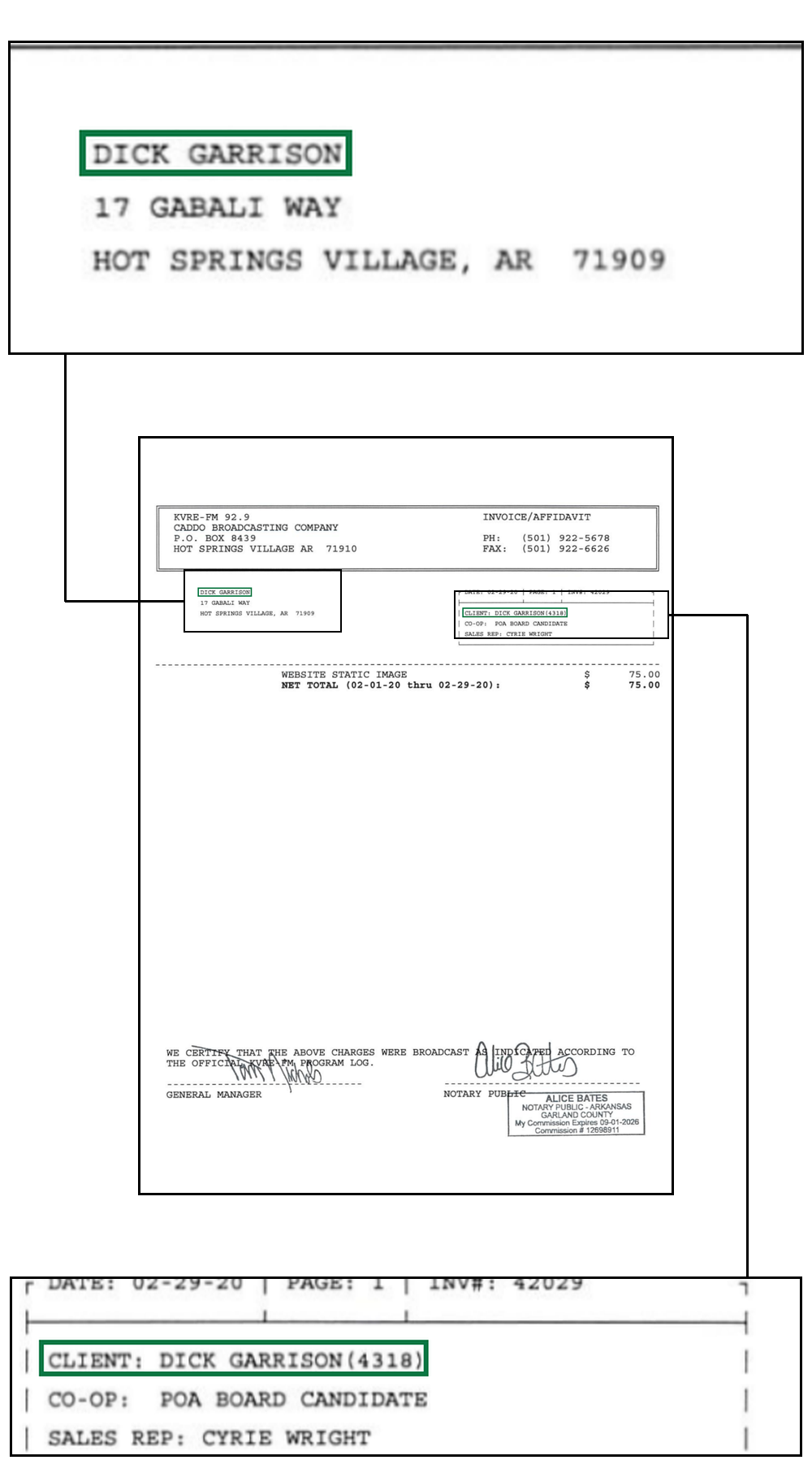}{DocILE\_697395}\hfill
\casepanel{\textbf{Reference 1}\\ \textbf{Ground truth:} BRUCE WESTERMAN}{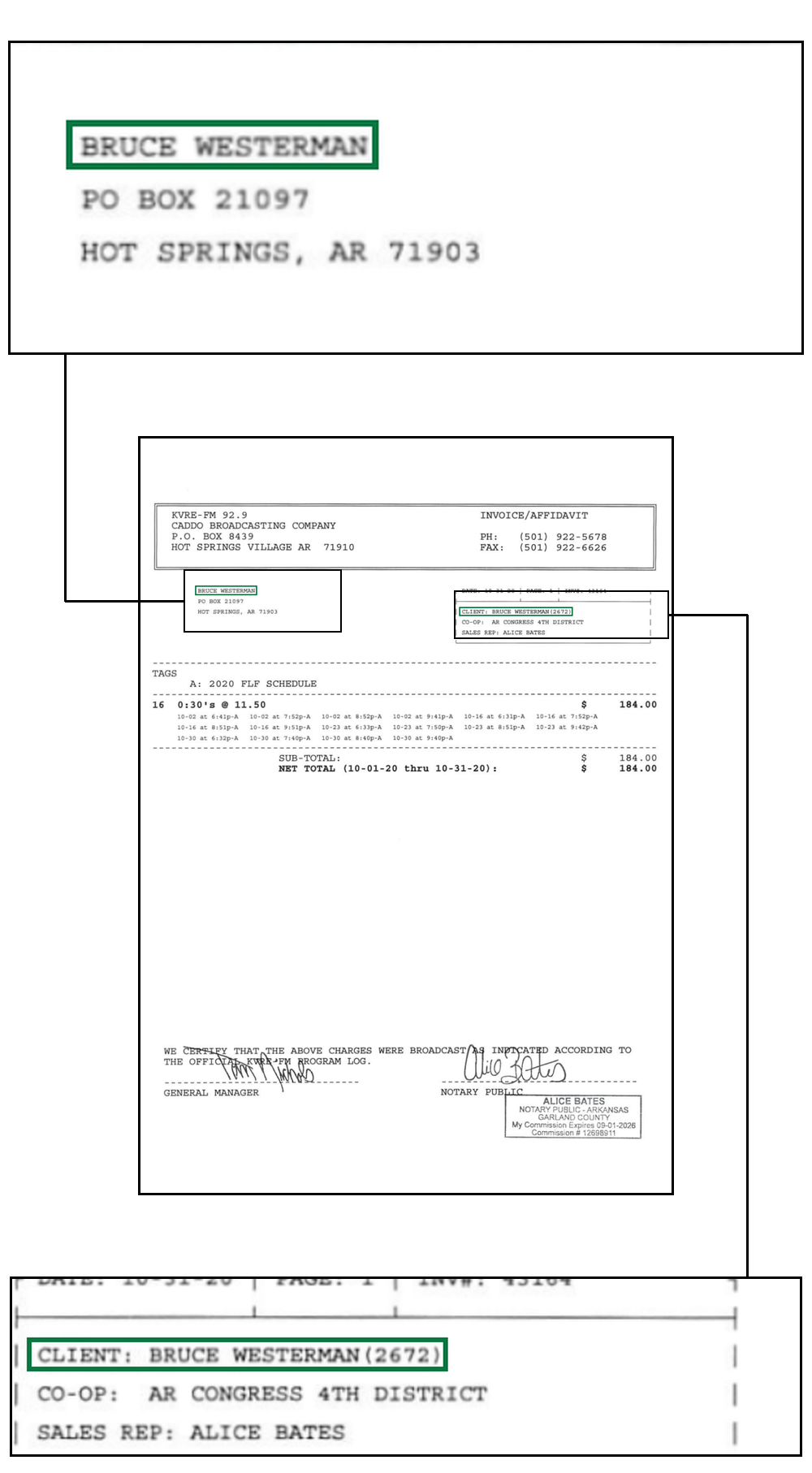}{DocILE\_629452}\hfill
\casepanel{\textbf{Reference 2}\\ \textbf{Ground truth:} \texttt{null}}{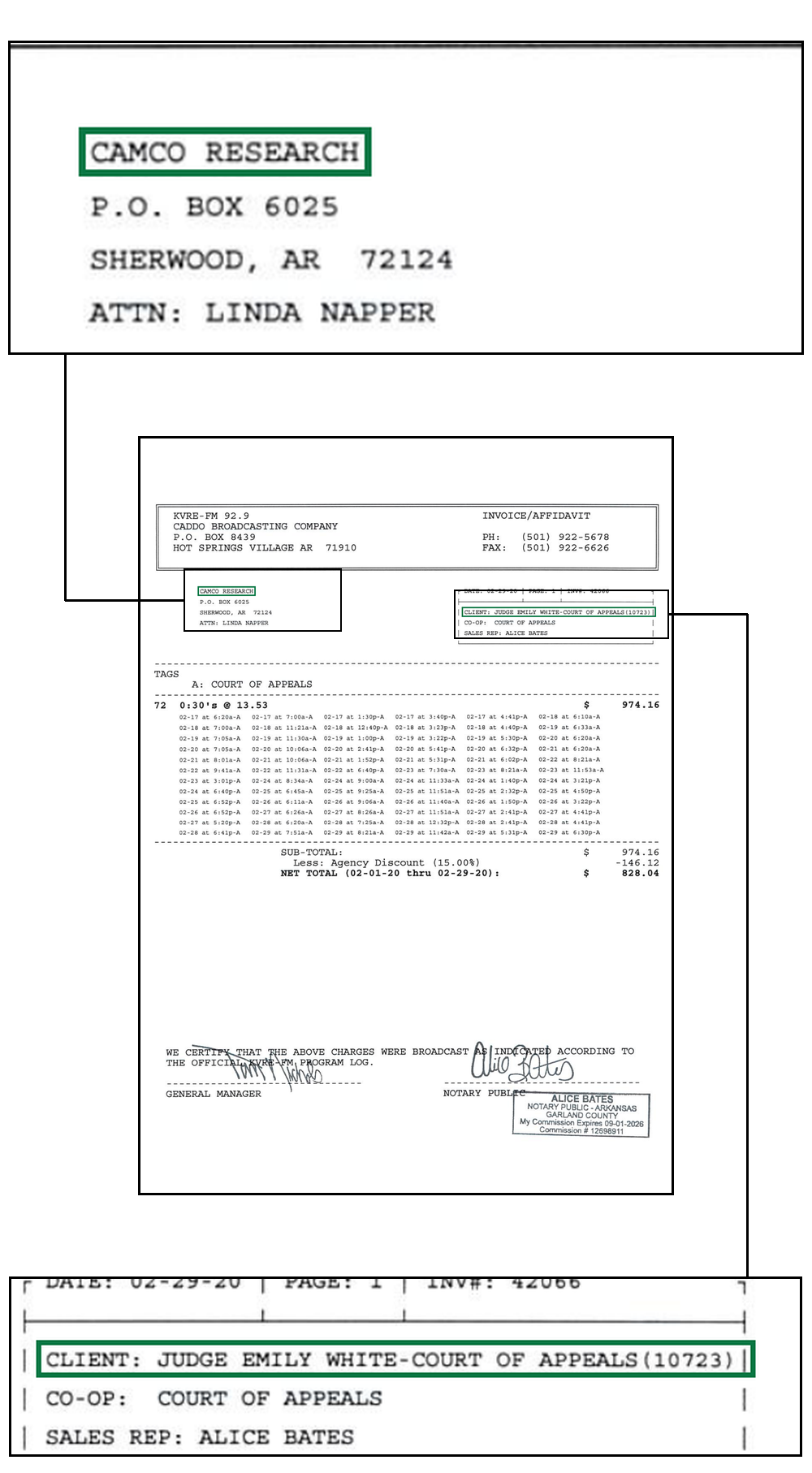}{DocILE\_349514}\hfill
\casepanel{\textbf{EvoKIE target}\\ \textbf{Prediction:} DICK GARRISON}{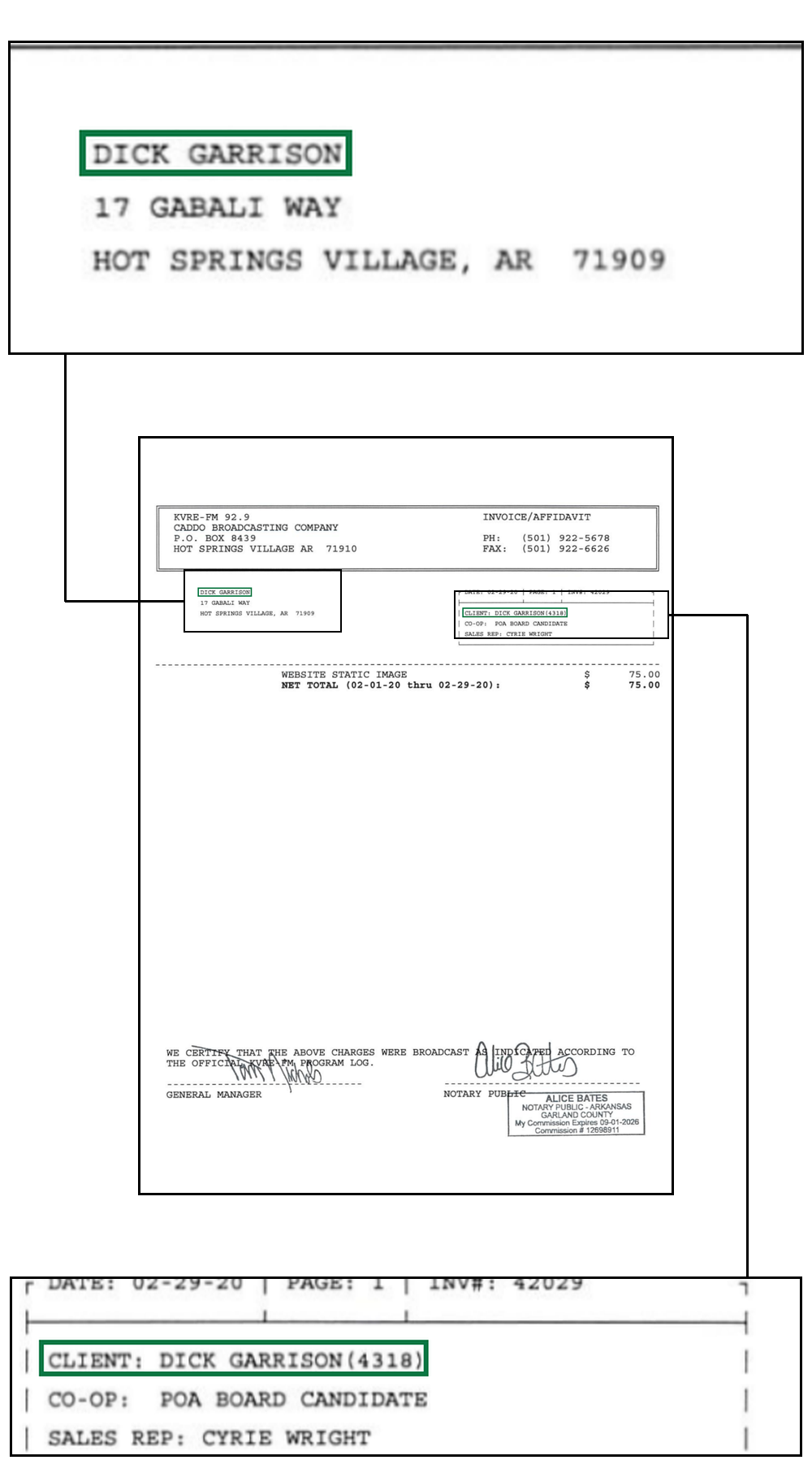}{DocILE\_697395}\par
\vspace{2pt}
\cardtitle{\textbf{(b)}\quad DocILE\_800733\quad Ground truth: \texttt{\{"customer\_other\_name": null\}}}
\casepanel{\textbf{SAIL target}\\ \textbf{Prediction:} SRCP VTB Liberal Judges}{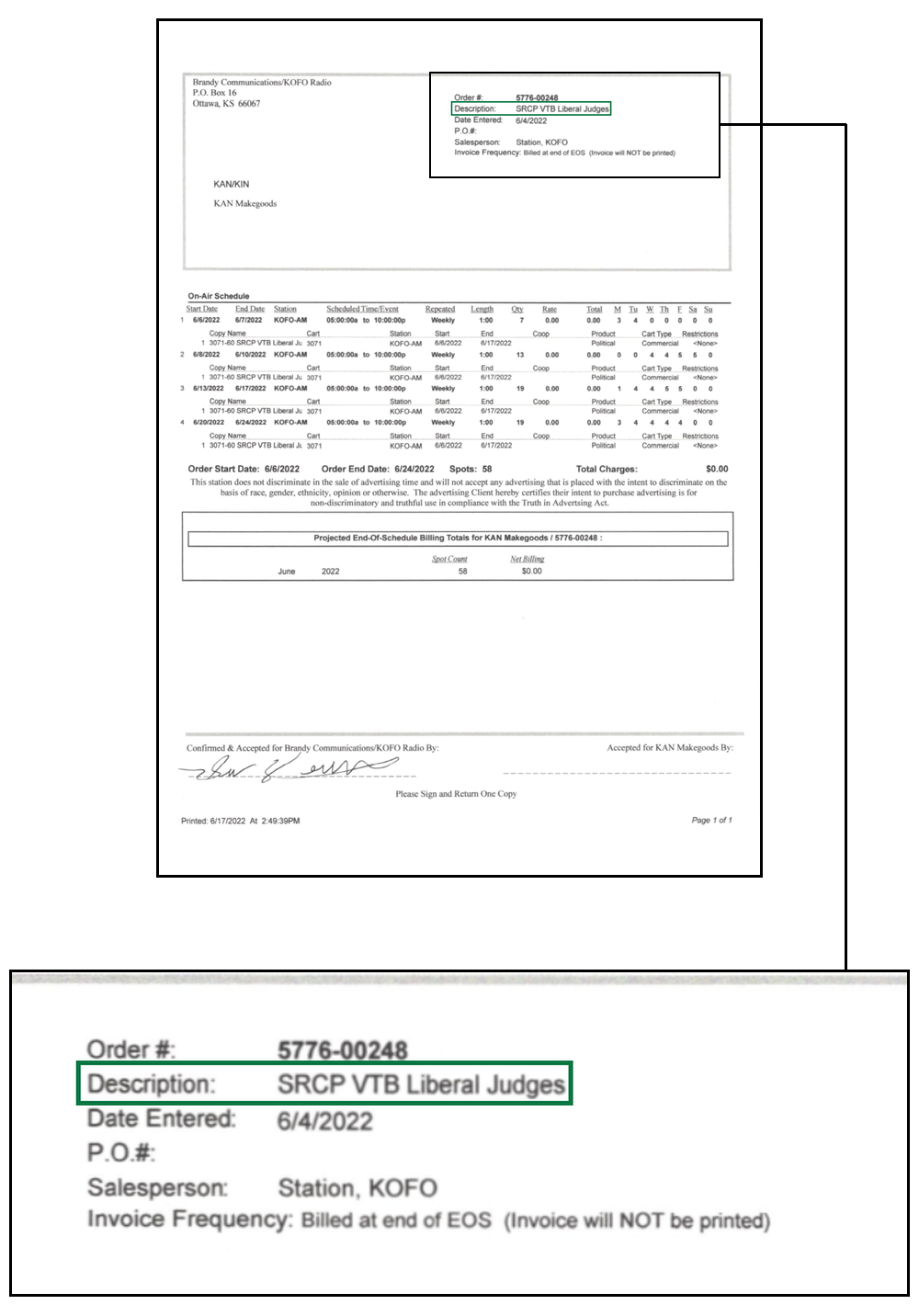}{DocILE\_800733}\hfill
\casepanel{\textbf{Reference 1}\\ \textbf{Ground truth:} \texttt{null}}{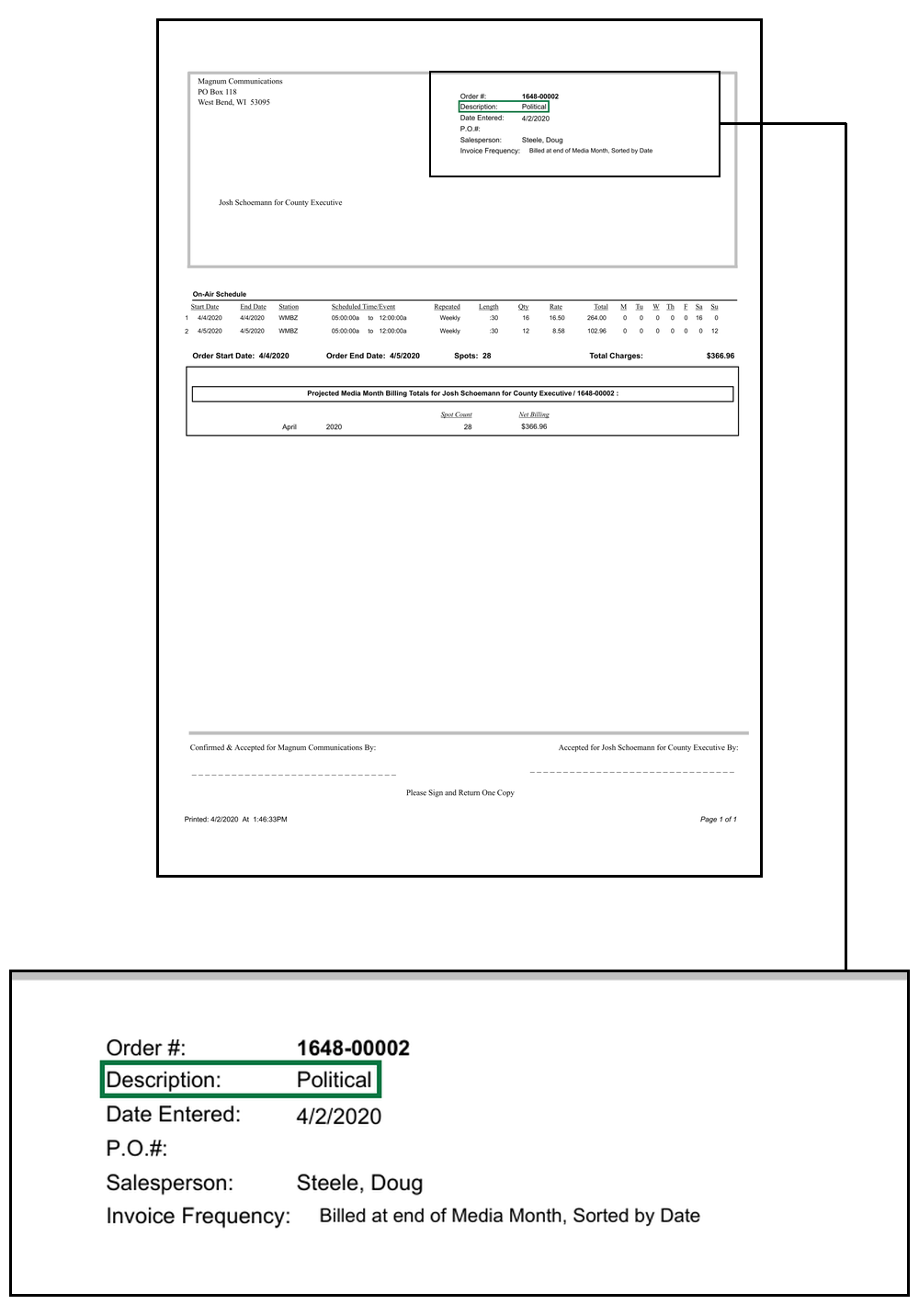}{DocILE\_747091}\hfill
\casepanel{\textbf{Reference 2}\\ \textbf{Ground truth:} GEORGIANS FOR KELLY LOEFFLER}{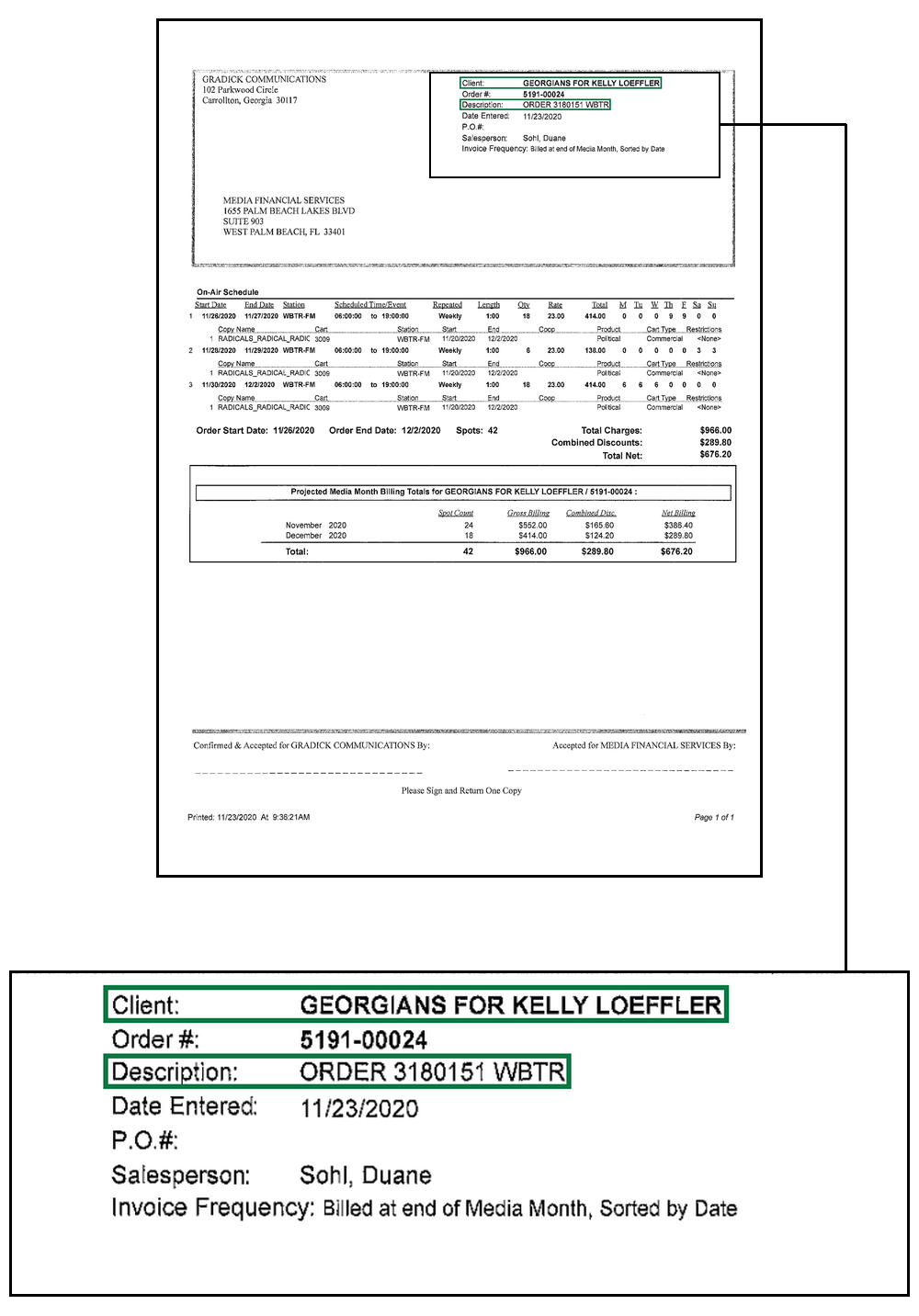}{DocILE\_137967}\hfill
\casepanel{\textbf{EvoKIE target}\\ \textbf{Prediction:} \texttt{null}}{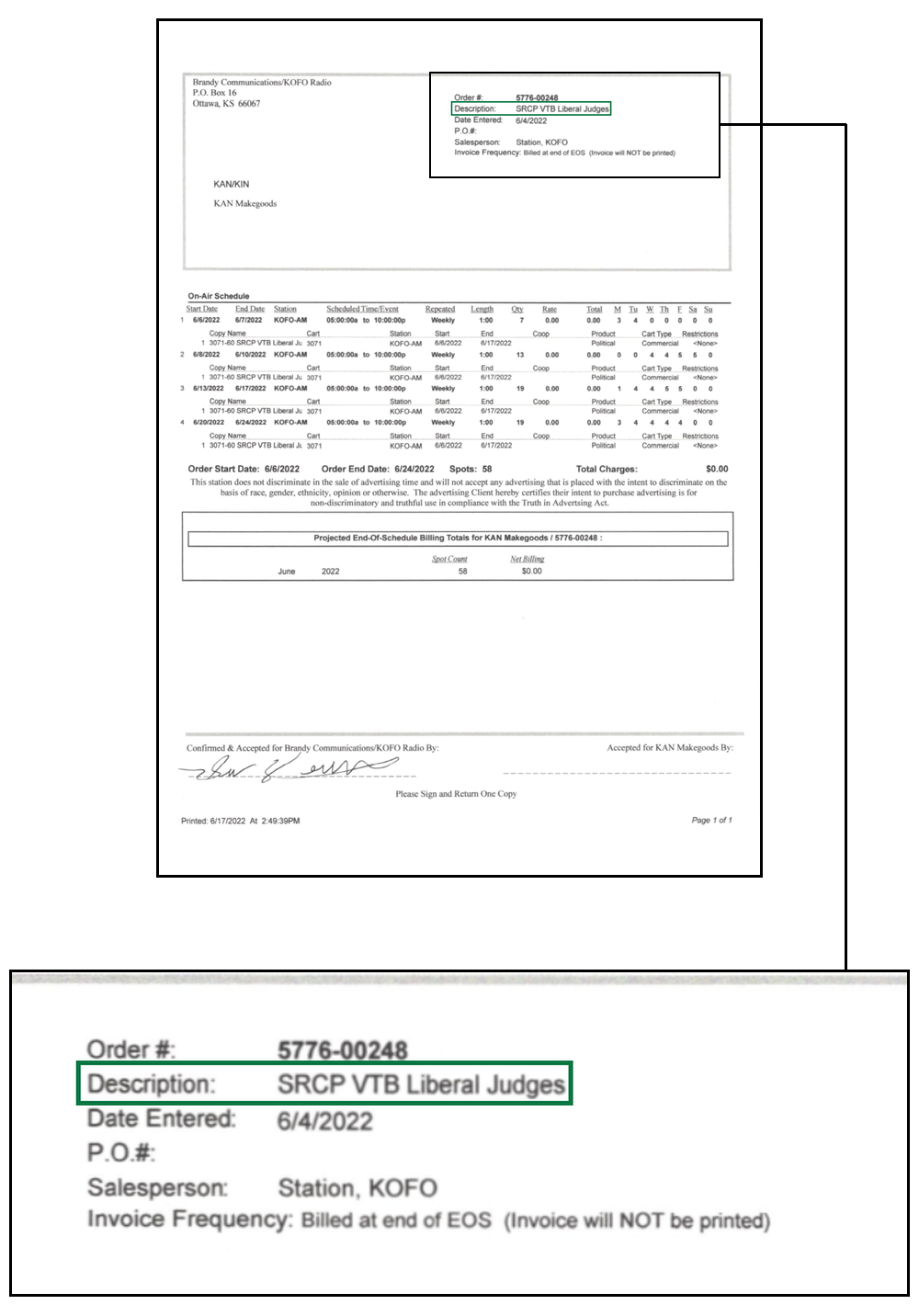}{DocILE\_800733}\par
\vspace{2pt}
\cardtitle{\textbf{(c)}\quad FATURA\_165435\quad Ground truth: \texttt{\{"AMOUNT\_DUE": "BALANCE\_DUE : 922.93 EUR"\}}}
\casepanel{\textbf{SAIL target}\\ \textbf{Prediction:} \texttt{null}}{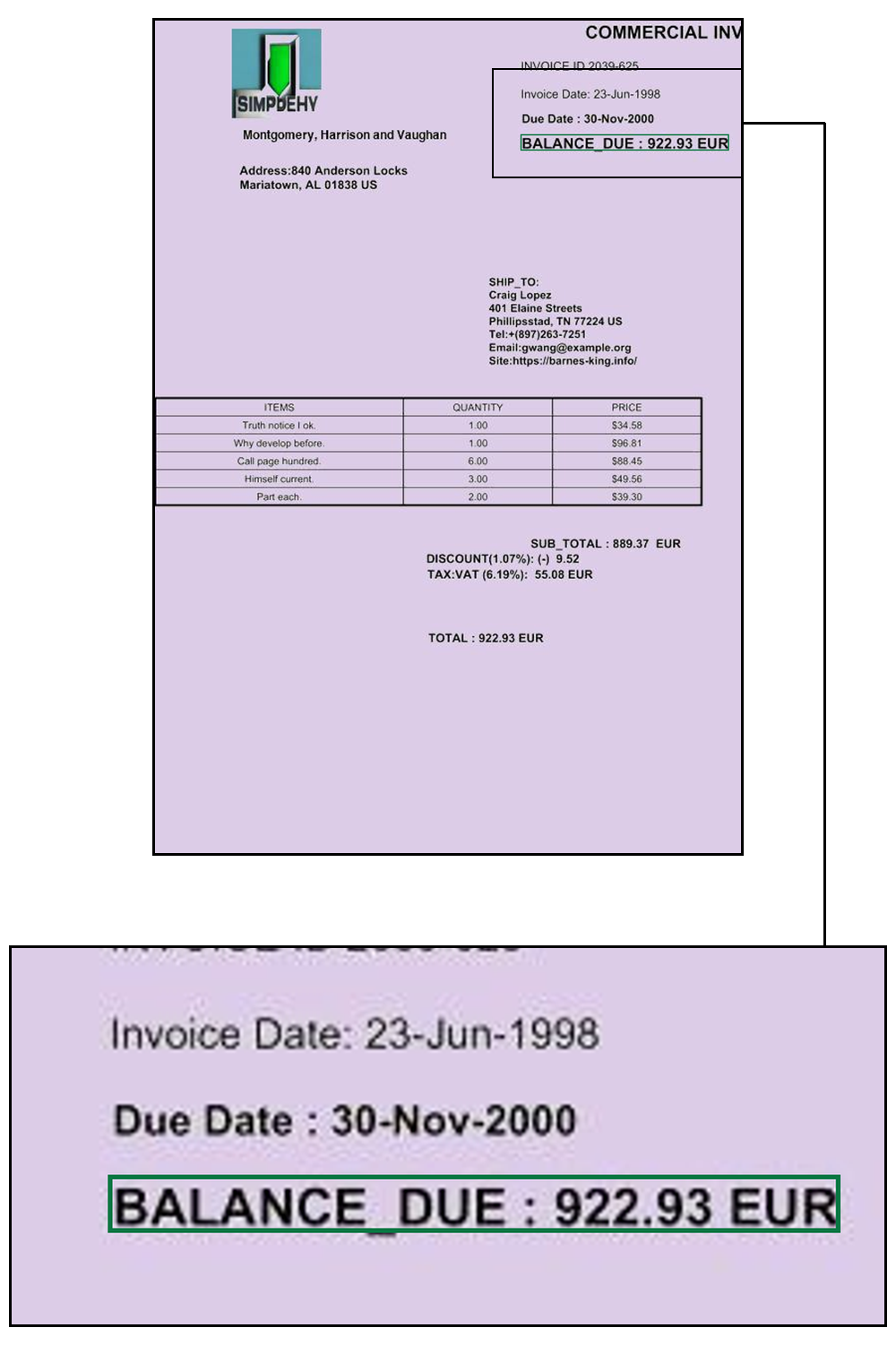}{FATURA\_165435}\hfill
\casepanel{\textbf{Reference 1}\\ \textbf{Ground truth:} AMOUNT\_DUE : 284.13 \$}{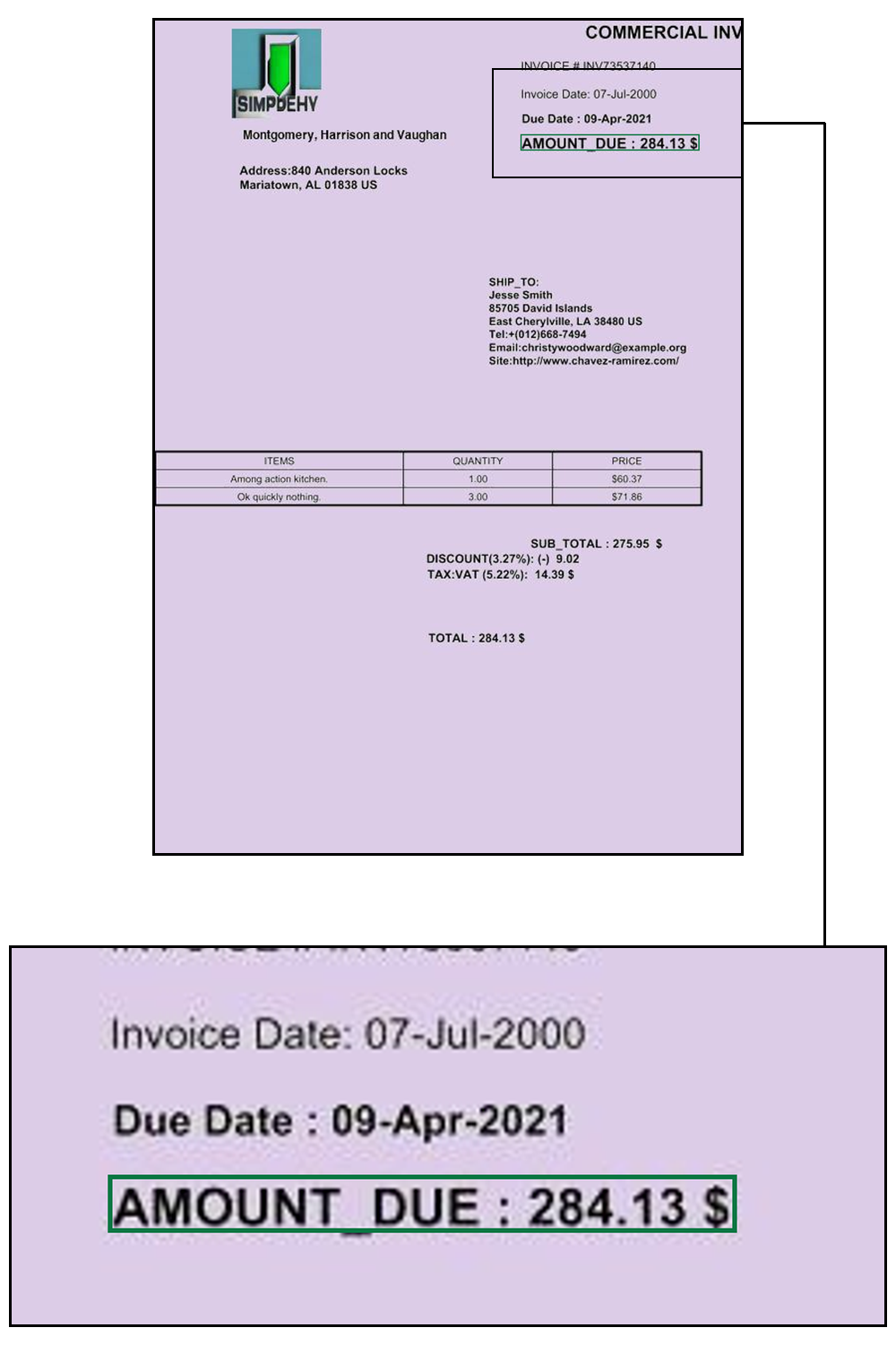}{FATURA\_758513}\hfill
\casepanel{\textbf{Reference 2}\\ \textbf{Ground truth:} DUE\_AMOUNT : 404.66 \$}{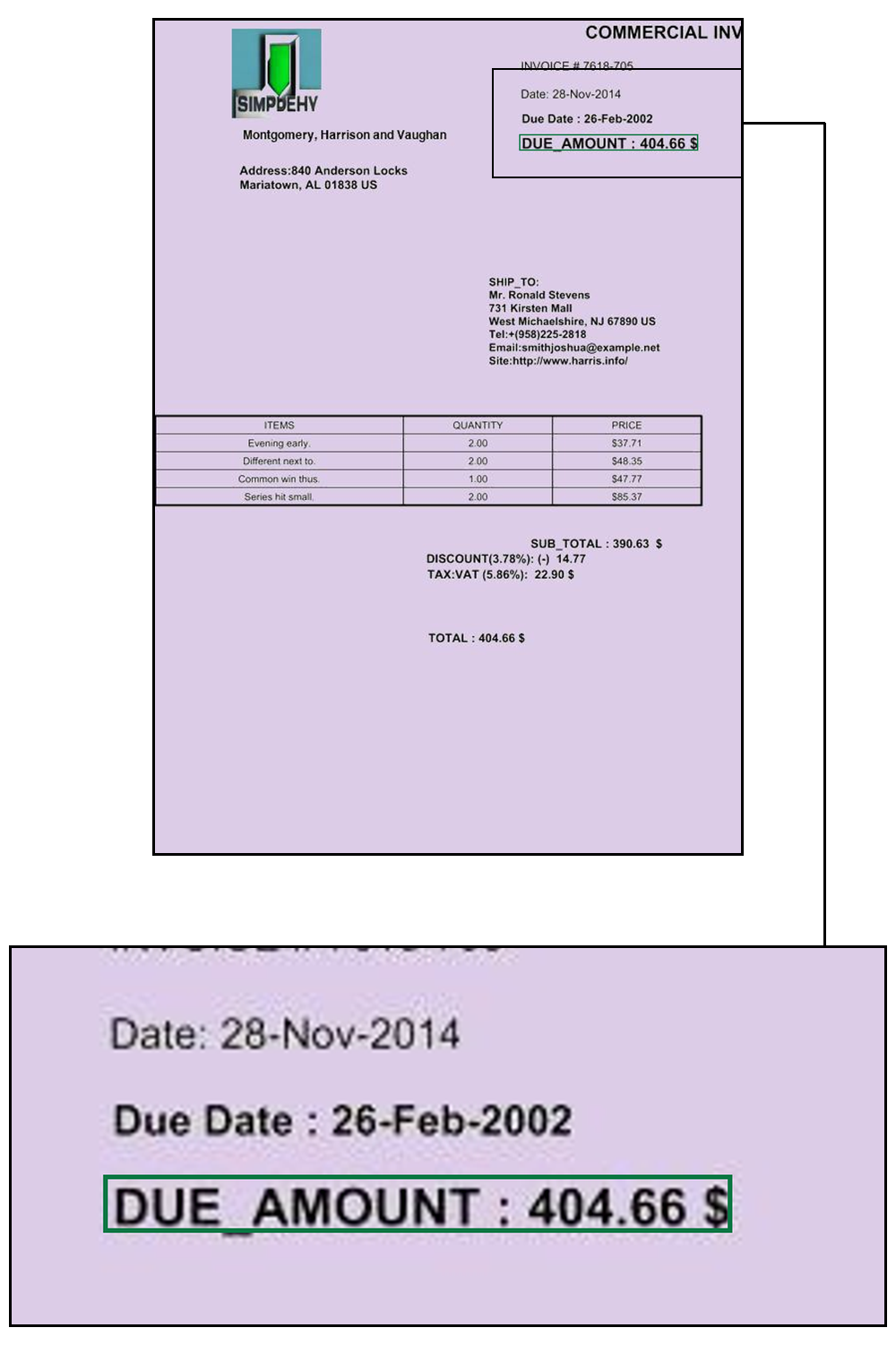}{FATURA\_505531}\hfill
\casepanel{\textbf{EvoKIE target}\\ \textbf{Prediction:} BALANCE\_DUE : 922.93 EUR}{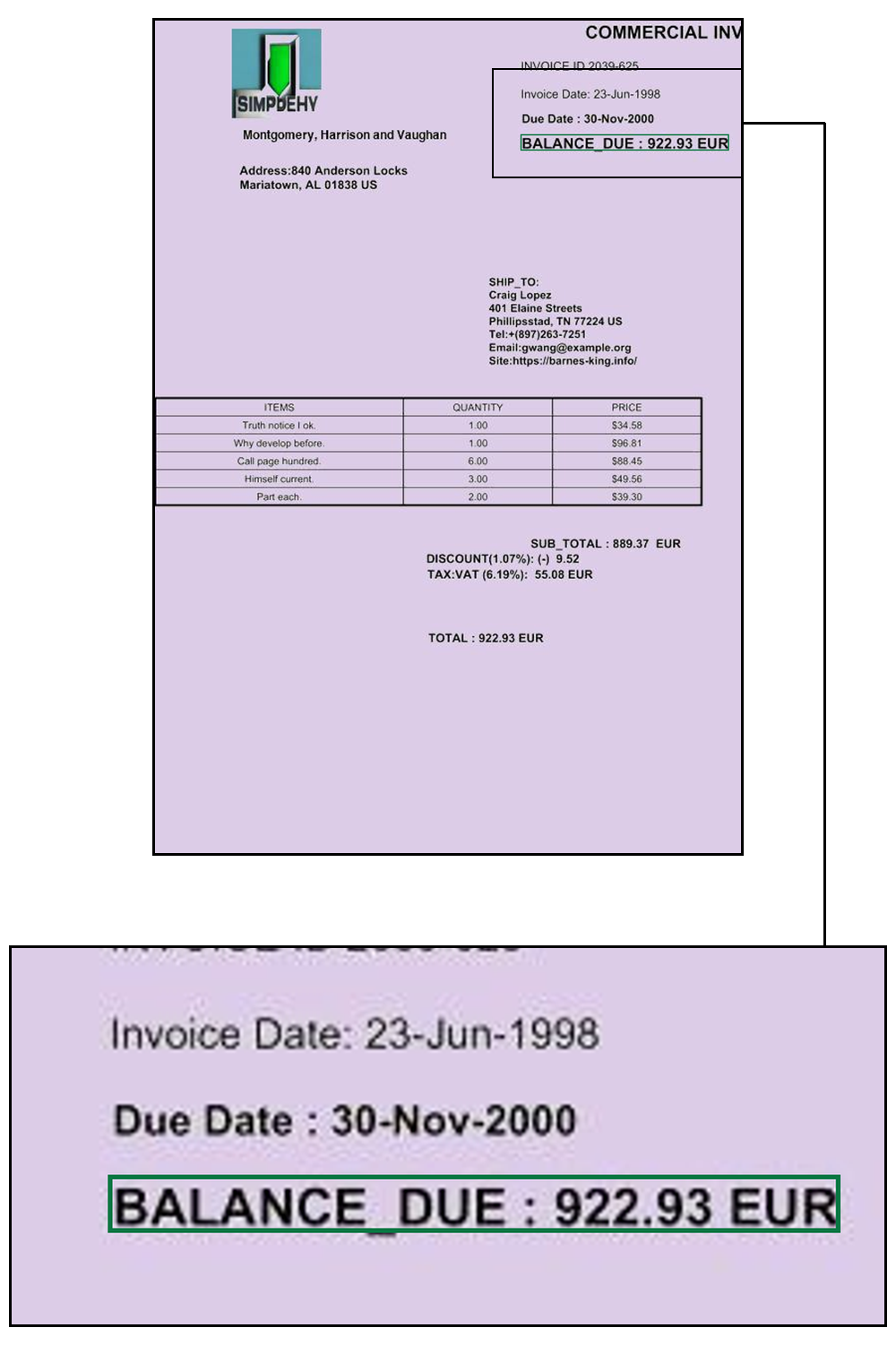}{FATURA\_165435}\par
\par\vspace{2pt}{\centering\fontsize{8.5}{10}\selectfont\legendbox{EvidenceGreen}{Evidence}\par}
\legenddivider
\caption{\textbf{Extraction-rule variation in retrieved examples and corresponding EvoKIE outputs.} Each panel shows SAIL's output on the target, two retrieved references, and EvoKIE's output on the same target, from left to right.
\textbf{(a)} The references require different outputs depending on whether the billed-customer and CLIENT names match. SAIL returns null despite matching names in the target; EvoKIE applies the matching-name condition.
\textbf{(b)} The references differ in whether an independent CLIENT entry is present. SAIL incorrectly extracts Description; EvoKIE returns null when no independent CLIENT entry is present.
\textbf{(c)} SAIL fails to retrieve available examples sharing the target's BALANCE\_DUE anchor, instead retrieving references with AMOUNT\_DUE and DUE\_AMOUNT; EvoKIE handles all three label variants.}
\label{fig:grouped-sail}
\end{figure}
\clearpage
\subsection{Typical Failure Cases of LRSyn}
LRSyn~\cite{parthasarathy2022landmarks} initially groups documents by global layout similarity, then merges groups based on similarities between local regions around field landmarks to synthesize shared extraction programs. However, local structural similarity does not necessarily imply extraction-rule compatibility. Figure~\ref{fig:grouped-lrsyn} illustrates two limitations: multiple rules may match the target but produce different outputs (a), and a shared rule may fail when label--value relationships or field semantics differ across documents (b,c). Through feedback-driven regrouping, EvoKIE jointly refines document grouping and extraction programs to support program reuse across documents with compatible extraction requirements. Router optimization adapts routing decisions to the revised groups, helping direct incoming documents to applicable programs and reducing incorrect assignments to programs with incompatible extraction rules.
\begin{figure}[H]
\renewcommand{\panelheight}{1.18in}
\cardtitle{\textbf{(a)}\quad Oversea\_697395\quad GT: \texttt{\{"taxAmount": "103,031.00"\}}}
\casepanel{\textbf{LRSyn target}\\ \textbf{Prediction:} 2,060,619.00}{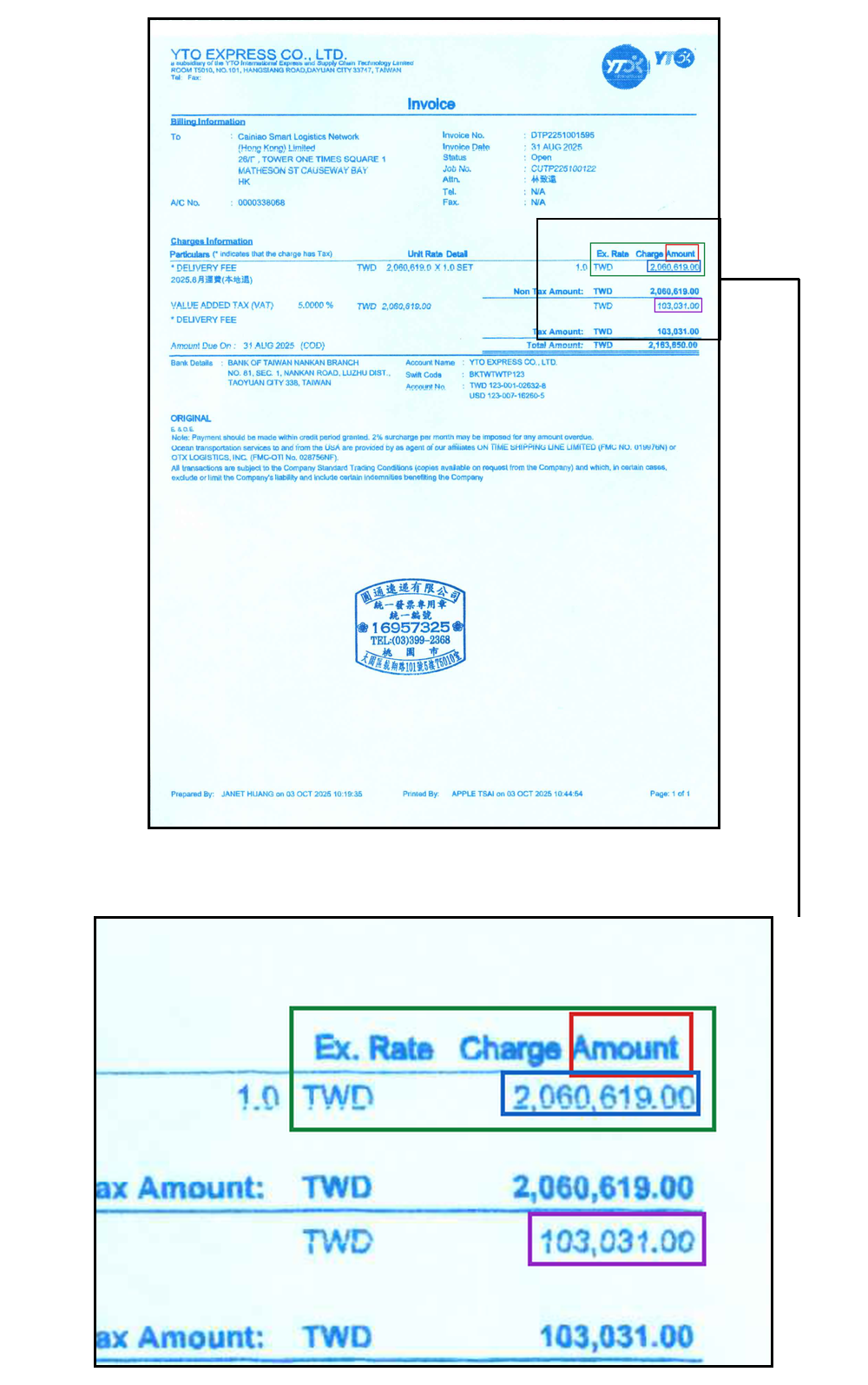}{Oversea\_697395}\hfill
\casepanel{\textbf{Rule 7}\\ \textbf{GT:} 103,031.00\par\textbf{Rule output:} 2,060,619.00}{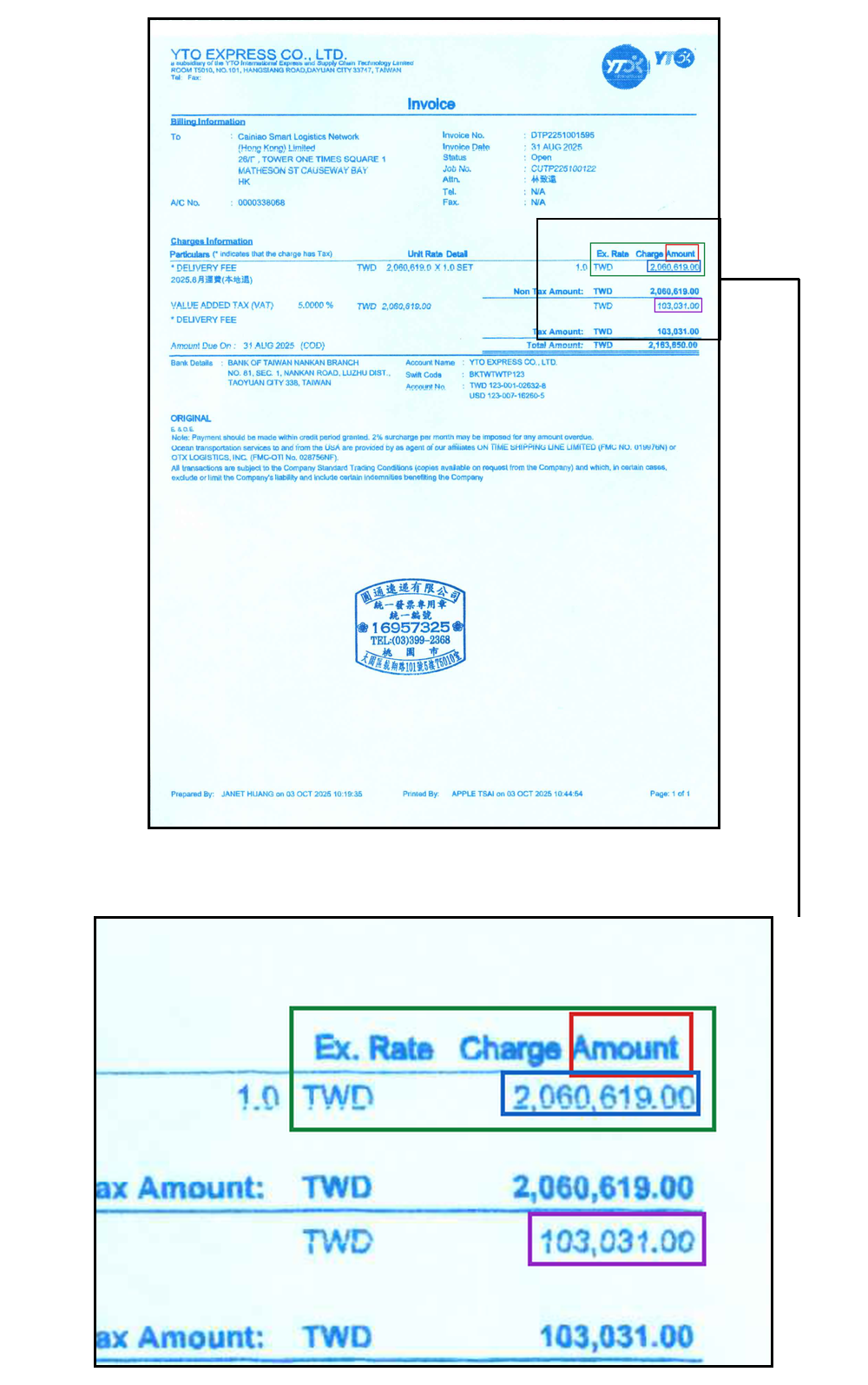}{Oversea\_697395}\hfill
\casepanel{\textbf{Rule 11}\\ \textbf{GT:} 103,031.00\par\textbf{Rule output:} 103,031.00}{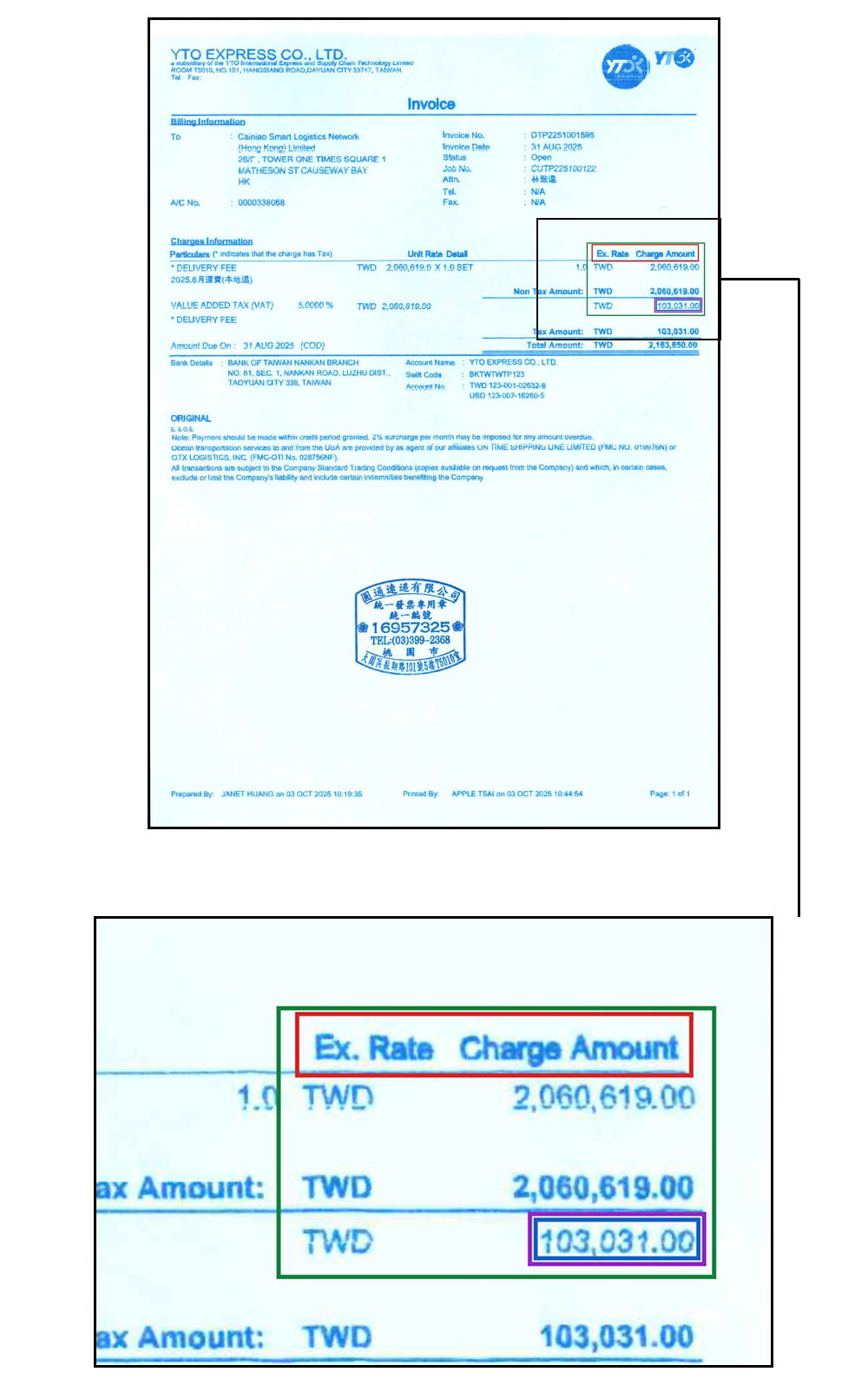}{Oversea\_697395}\hfill
\casepanel{\textbf{EvoKIE target}\\ \textbf{Prediction:} 103,031.00}{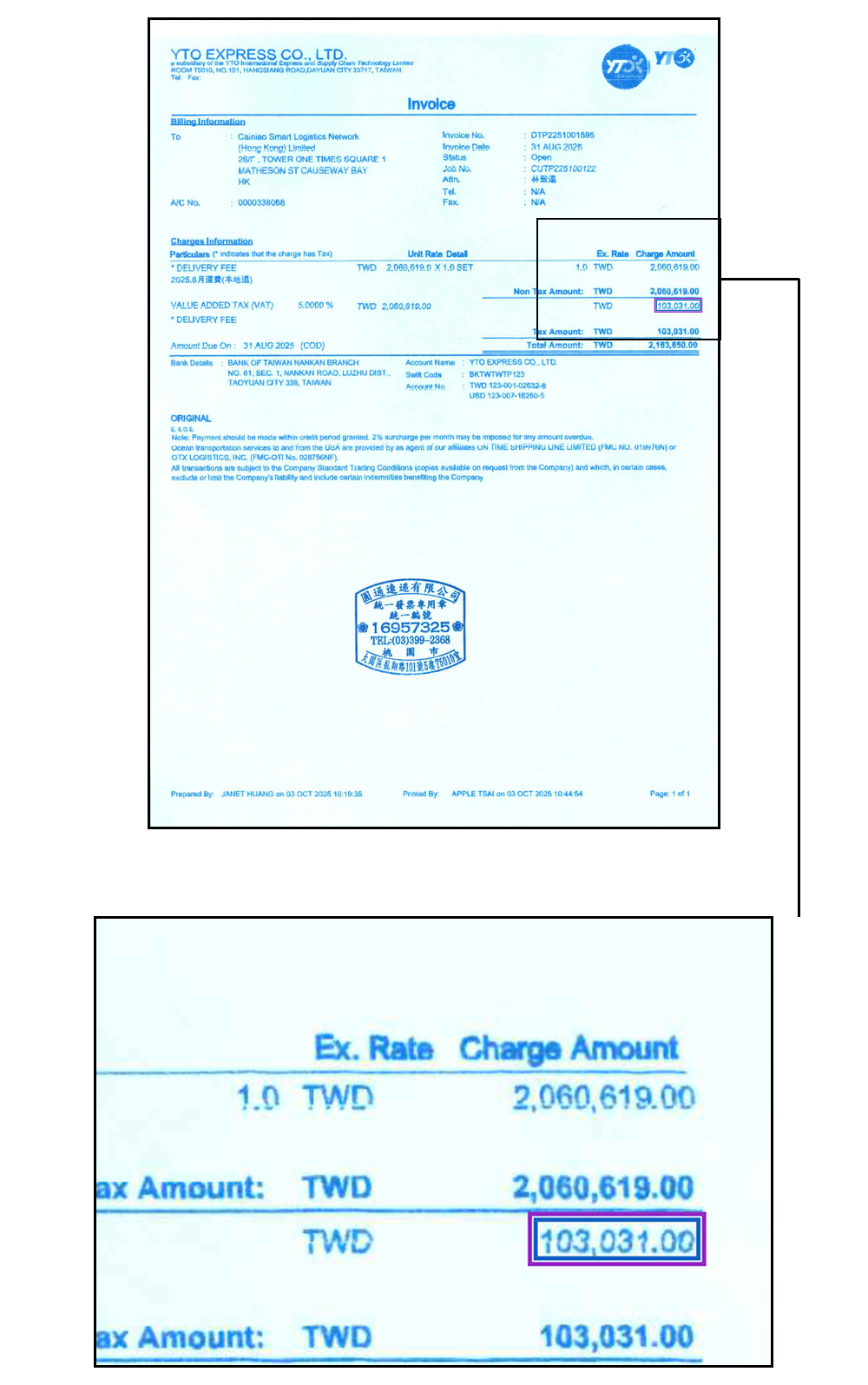}{Oversea\_697395}\par
\vspace{2pt}
\cardtitle{\textbf{(b)}\quad Oversea\_800733\quad GT: \texttt{\{"sellerIdentification": "12167452-2-44"\}}}
\casepanel{\textbf{LRSyn target}\\ \textbf{Prediction:} VAT}{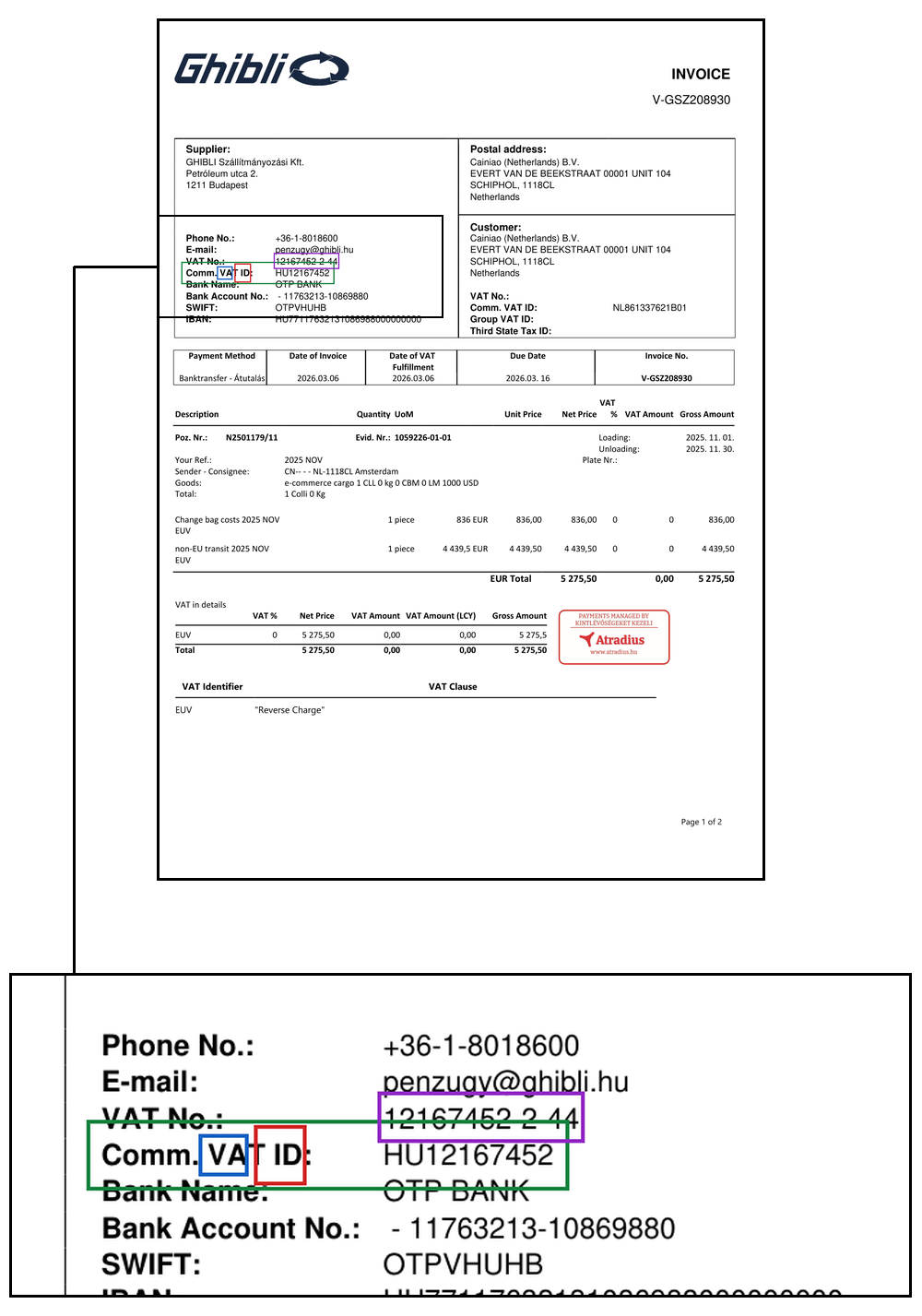}{Oversea\_800733}\hfill
\casepanel{\textbf{Reference}\\ \textbf{GT:} FR55464200369\par\textbf{Rule output:} FR55464200369}{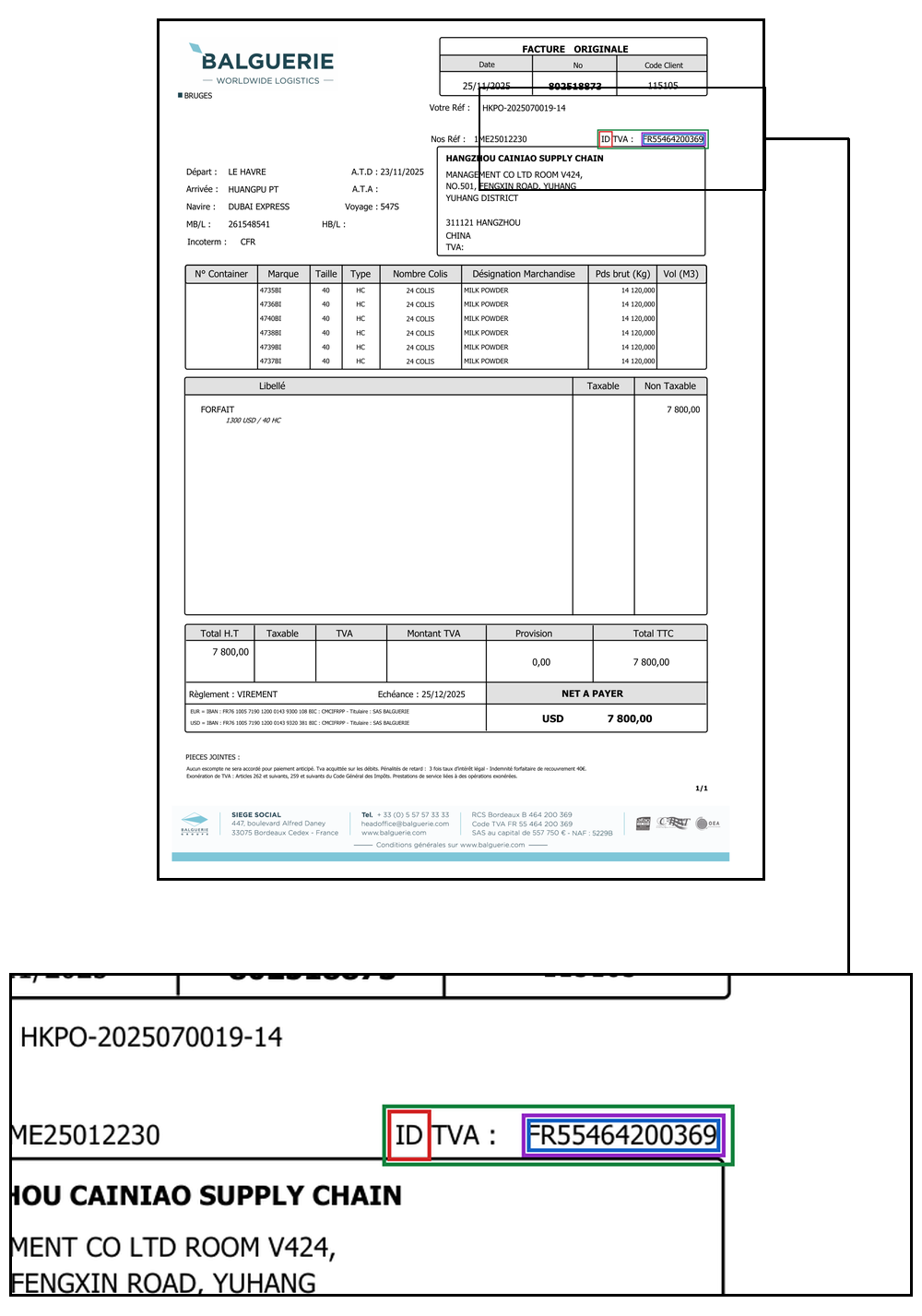}{Oversea\_891514}\hfill
\casepanel{\textbf{EvoKIE target}\\ \textbf{Prediction:} 12167452-2-44}{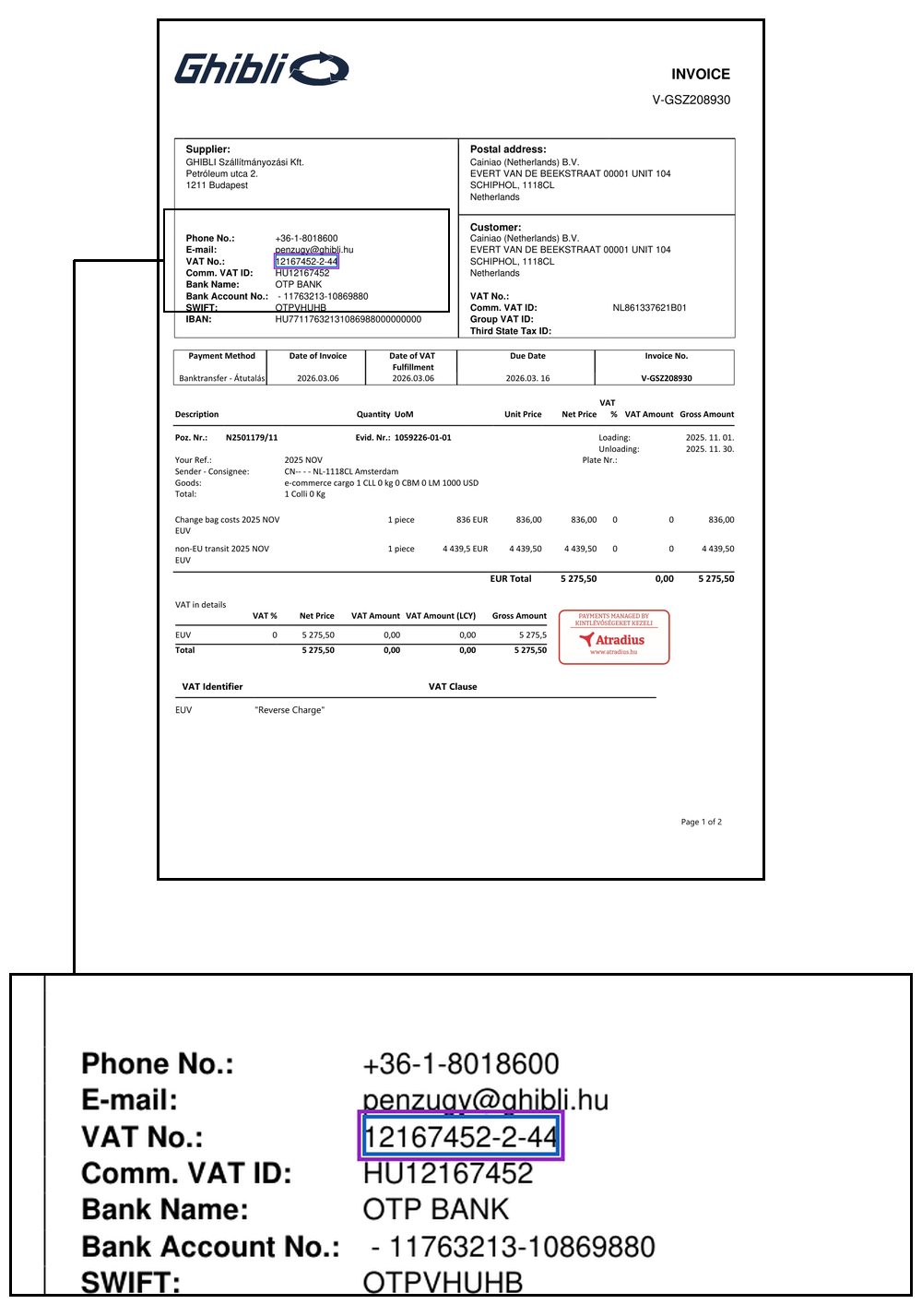}{Oversea\_800733}\par
\vspace{2pt}
\cardtitle{\textbf{(c)}\quad FATURA\_137967\quad GT: \texttt{\{"TOTAL": "TOTAL : 728.05 USD"\}}}
\casepanel{\textbf{LRSyn target}\\ \textbf{Prediction:} Total in words: seven hundred and twent-}{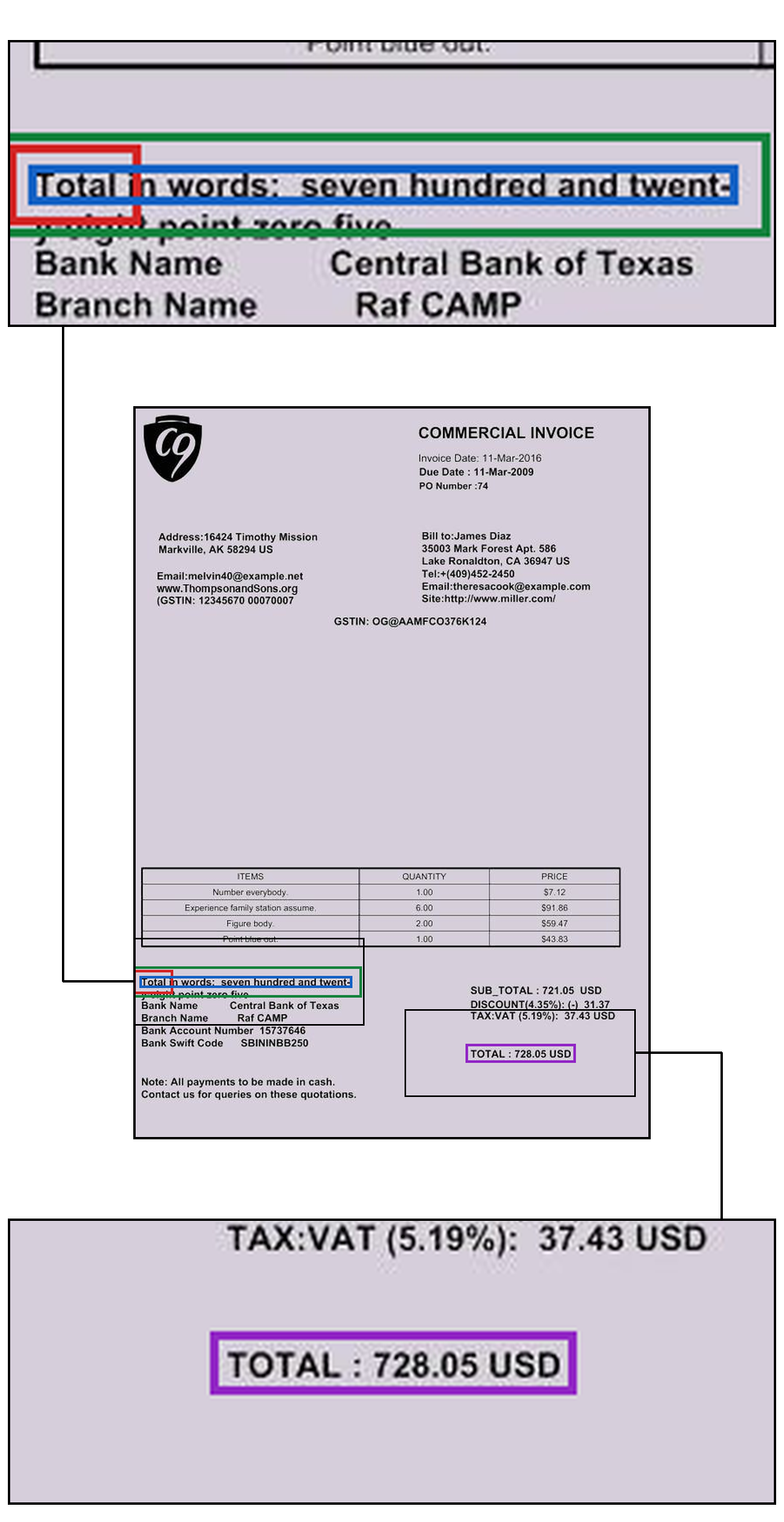}{FATURA\_137967}\hfill
\casepanel{\textbf{Reference}\\ \textbf{GT:} TOTAL : 141.88 USD\par\textbf{Rule output:} TOTAL: 141.88 USD}{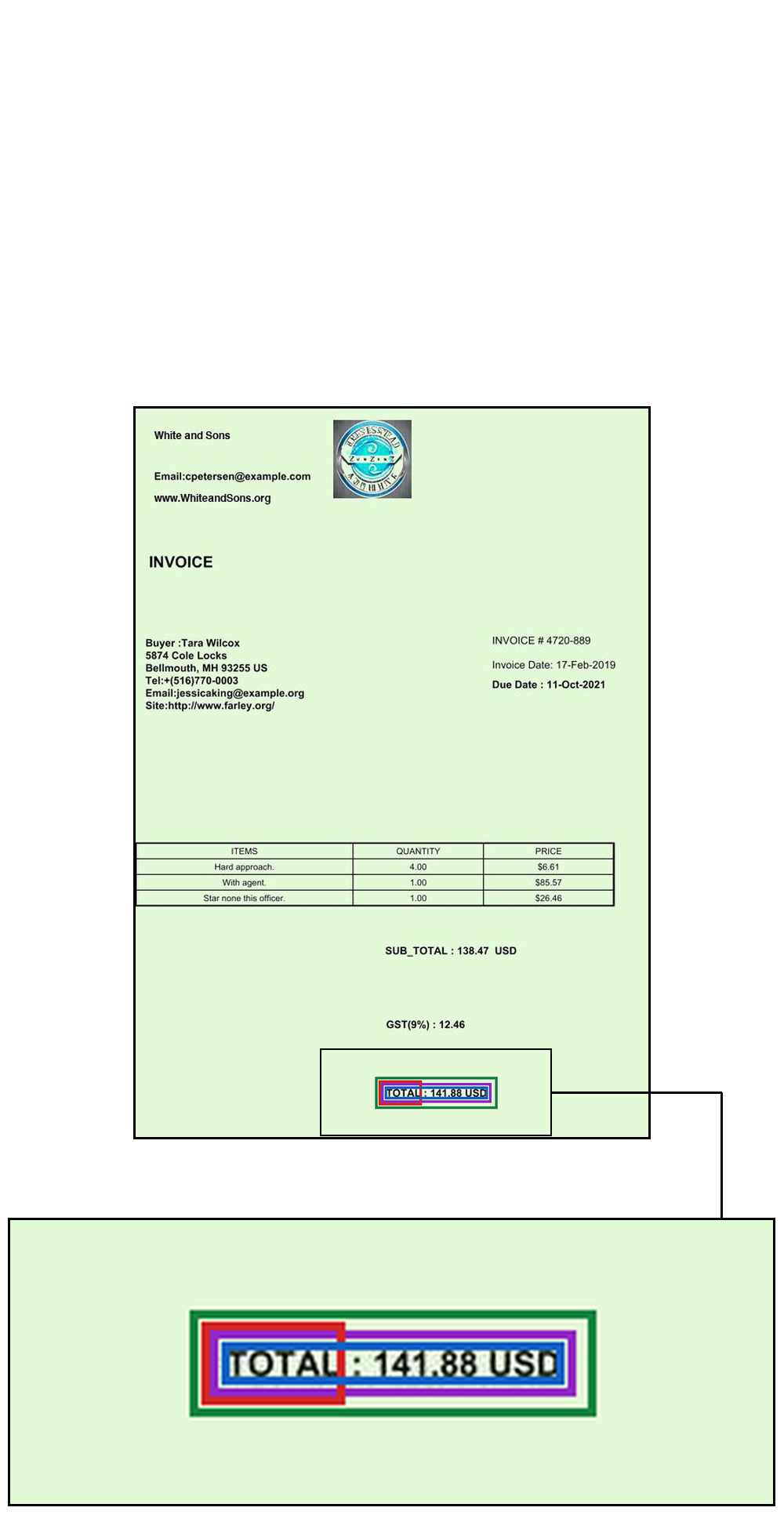}{FATURA\_703693}\hfill
\casepanel{\textbf{EvoKIE target}\\ \textbf{Prediction:} TOTAL : 728.05 USD}{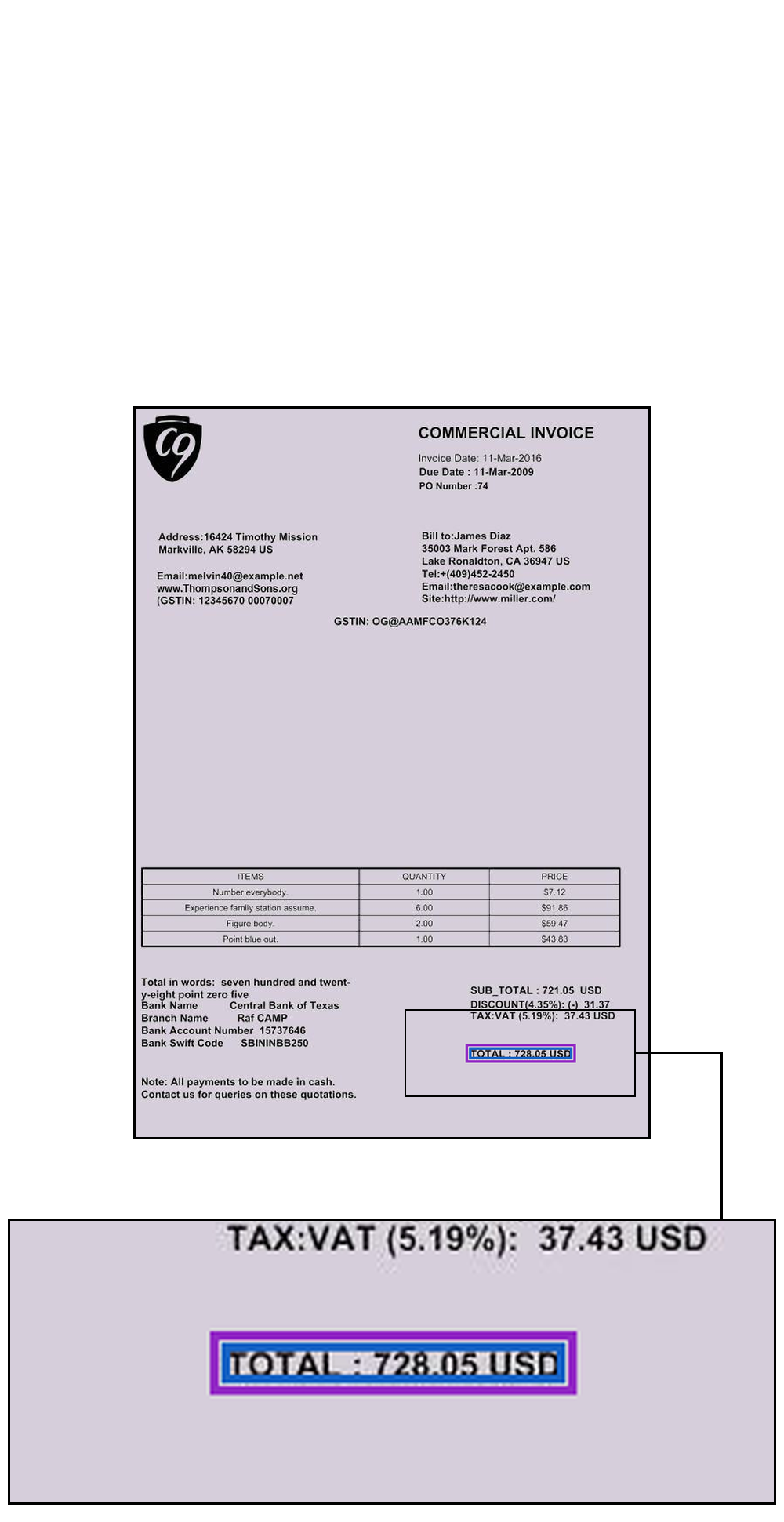}{FATURA\_137967}\par
\par\vspace{2pt}{\centering\fontsize{8.5}{10}\selectfont\legendbox{LandmarkRed}{Landmark}\quad\legendbox{RegionGreen}{Local region}\quad\legendbox{PredBlue}{Prediction}\quad\legendbox{GTPurple}{Ground truth}\par}
\legenddivider
\caption{\textbf{Candidate-rule ambiguity and shared-rule failures in LRSyn, with corresponding EvoKIE outputs.} Each panel shows LRSyn's output on the target, diagnostic rule or reference examples, and EvoKIE's output on the same target, from left to right.
\textbf{(a)} Two accepted rules select the charge and tax amounts, respectively. LRSyn selects the first nonempty output, missing the correct but unselected candidate; EvoKIE selects the rightmost amount beside Tax Amount.
\textbf{(b)} The ID-based rule succeeds on its reference but selects the label VAT on the target. EvoKIE uses tax-ID formats and seller/buyer context to identify the seller's tax ID.
\textbf{(c)} The generic Total landmark matches Total in words. EvoKIE matches TOTAL: or TOTAL : and returns the numeric-total text.}
\label{fig:grouped-lrsyn}
\end{figure}
\endgroup

\clearpage
\section{Analysis of the Co-Evolution Process}
\label{sec:supp-co-evolution-analysis}
\subsection{Global Evolution of Document Grouping}
\begin{figure}[p]
\centering
\includegraphics[width=\textwidth,height=.84\textheight,keepaspectratio]{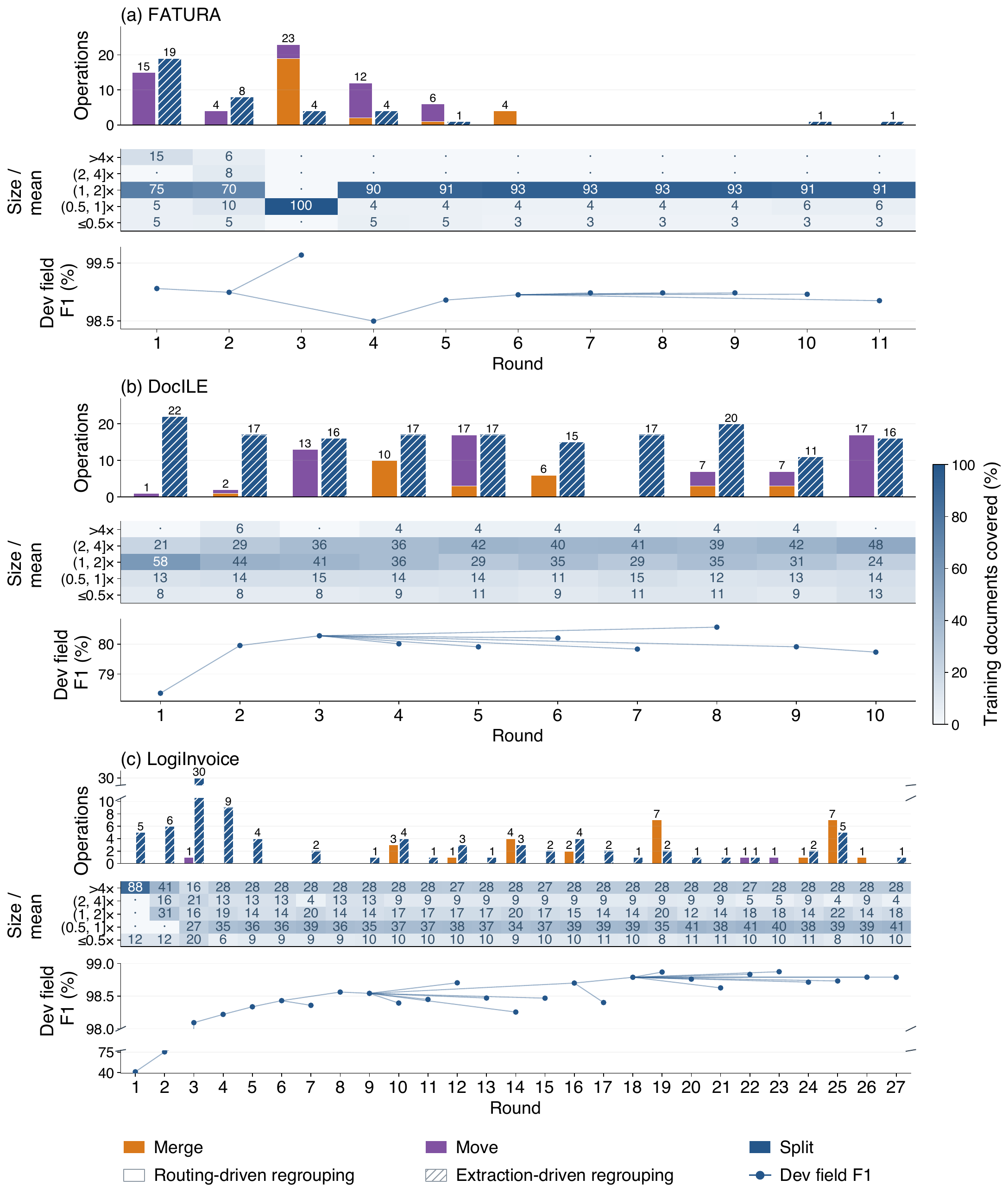}
\caption{Global evolution of document grouping across co-evolution rounds on (a) FATURA, (b) DocILE, and (c) LogiInvoice. Top: regrouping operation counts, with orange, purple, and blue denoting merging, moving, and splitting, respectively. Solid and hatched bars indicate routing-guided and extraction-guided regrouping. Middle: group-size distributions normalized by the current mean group size; cell values show the percentage of training documents belonging to groups within each size interval. Bottom: development-set field F1, with edges connecting actual parent--child versions.}
\label{fig:global-evolution}
\end{figure}

To characterize grouping evolution across co-evolution rounds, Figure~\ref{fig:global-evolution} plots regrouping operation counts, group-size distributions, and development-set field F1 for each dataset. Each round applies move and merge operations during routing-guided regrouping, followed by split operations during extraction-guided regrouping. Each merge absorbs an entire source group into another group and counts as one operation; document reassignments outside merges each count as one move. Each split counts once per newly created nonempty subgroup, excluding the retained original group.

For each group $g \in \Pi_c$ in the partition at the end of round \(c\), we normalize its size by the mean group size:
\begin{equation}
    s_{c,g} = \frac{|g|}{N/G_c},
    \qquad G_c = |\Pi_c|,
\end{equation}
where $N$ is the number of training documents and $G_c$ is the number of groups in $\Pi_c$. The heatmap shows the percentage of training documents belonging to groups within each normalized size interval.

\textbf{Regrouping Patterns Across Rounds and Datasets.}
Regrouping follows dataset-specific trajectories, with operation types and update frequency changing across rounds. FATURA combines splitting and document reassignment in early rounds, shifts toward merging in intermediate rounds, and makes few partition updates in later rounds. DocILE uses splitting throughout, alongside recurrent document reassignment and merging. LogiInvoice starts with extensive splitting, followed by smaller updates involving all three operations.

\textbf{Evolution of Group-Size Distributions.}
Training documents occupy distinct relative group-size ranges across datasets. In later FATURA versions, approximately 91--93\% of training documents belong to groups between one and two times the current mean size. On DocILE, the share of training documents in groups two to four times the current mean size rises from 21\% in the first round to 48\% in the last. LogiInvoice has substantial document coverage in both below-mean and above-mean groups, with less coverage in between.

\subsection{Routing and Extraction Compatibility Across Rounds}
\begin{figure}[p]
\centering
\includegraphics[width=\textwidth,height=.84\textheight,keepaspectratio]{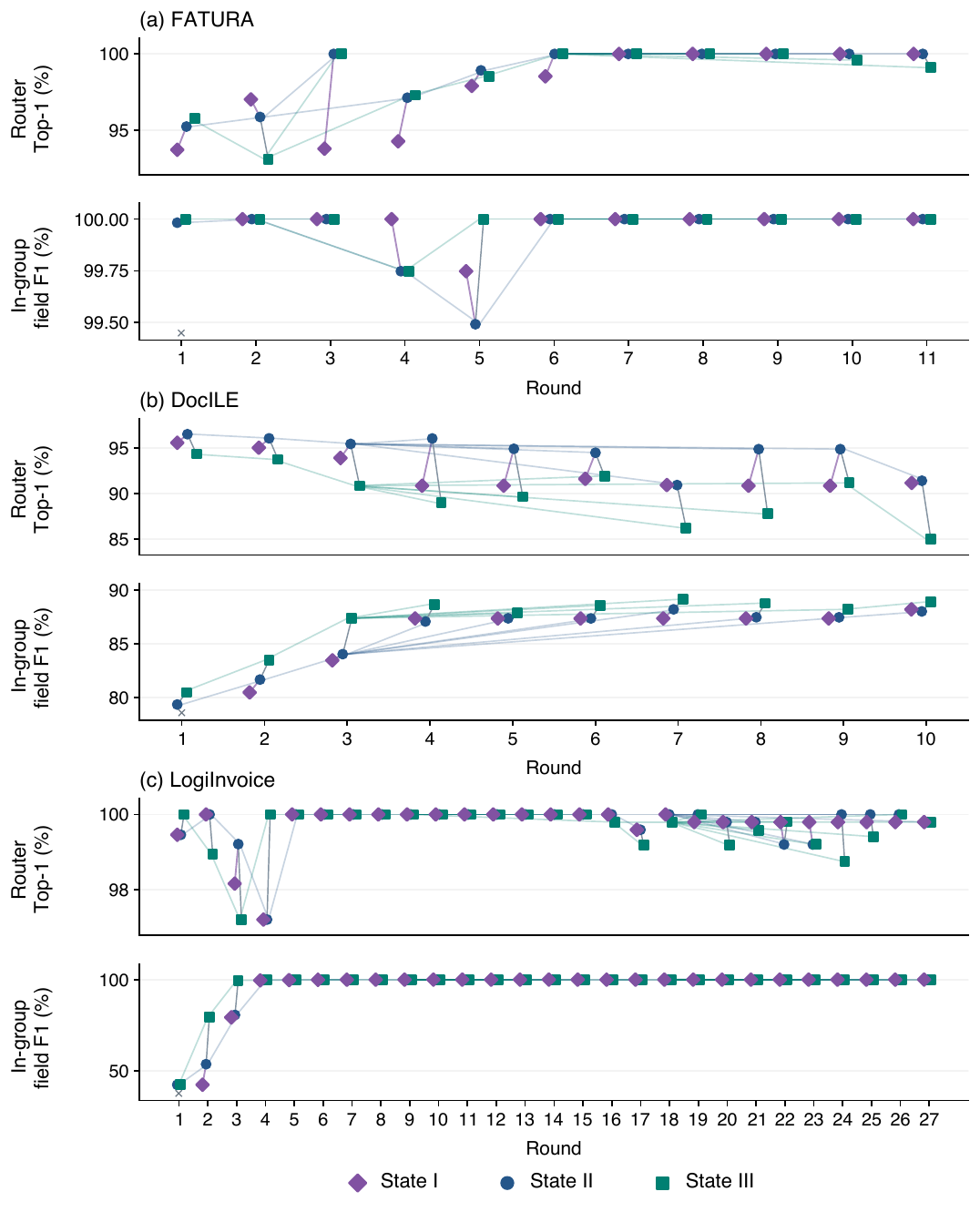}
\caption{Routing and extraction compatibility across co-evolution rounds on (a) FATURA, (b) DocILE, and (c) LogiInvoice. Top: held-out Top-1 routing accuracy. Bottom: in-group field F1 evaluated independently of routing. State I (purple diamonds) is after router optimization under the parent partition, with extraction programs inherited from the parent version. States II (blue circles) and III (teal squares) are after routing-guided and extraction-guided regrouping, respectively, including routing-index reconstruction and extraction-program updates. Within-round segments connect successive states; across-round edges connect the same state along actual parent--child versions.
}
\label{fig:module-adaptation}
\end{figure}

To examine how document grouping shapes routing and extraction during co-evolution, Figure~\ref{fig:module-adaptation} tracks the performance of both modules before and after each regrouping stage and across rounds. Routing performance is measured by held-out Top-1 routing accuracy, averaged over ten random splits of the training set following Section 3.3. For extraction, we report in-group field F1 by applying each group-specific program directly to the training documents assigned to its group under the current partition, independently of routing.

We track three states in each co-evolution round. First, router optimization updates the router under the parent partition, while the grouping and extraction programs remain inherited from the parent version. Second, routing-guided regrouping revises the partition using routing feedback, after which the routing index is rebuilt and group-specific extraction programs are synthesized and optimized under the revised partition. Third, extraction-guided regrouping further refines the partition using program execution feedback, followed by another routing-index rebuild and extraction-program optimization.

\textbf{Trade-off Between Routing and Extraction Compatibility.} 
The results reveal a trade-off between extraction-rule compatibility and group distinguishability during co-evolution. On DocILE, extraction-guided regrouping increases in-group field F1 in all ten rounds, by an average of 1.34 percentage points, while held-out Top-1 routing accuracy decreases by 4.58 percentage points on average. This pattern is consistent with the tension motivating our framework: groups that better support shared extraction rules can be harder for the current router to distinguish.

\textbf{Module Adaptation to Evolving Document Partitions.}
Agentic module optimization adapts the router and group-specific extraction programs to evolving document partitions. When the partition refined by extraction feedback in the previous round becomes the current parent partition, router optimization addresses the resulting decline in routing performance. It improves held-out Top-1 routing accuracy in five of ten DocILE rounds and five of eleven FATURA rounds, while preserving accuracy in the remaining rounds. After each of the two regrouping stages, agentic program synthesis adapts shared extraction rules to the revised groups. Following routing-guided regrouping, mean in-group field F1 increases by 0.49 and 0.27 percentage points on LogiInvoice and DocILE, respectively, relative to the parent programs evaluated under the parent partition. Subsequent program updates following extraction-guided regrouping yield mean gains of 1.66 and 1.34 percentage points, respectively. Together, these observations illustrate how module optimization supports routing and extraction under evolving group boundaries, contributing to an overall upward trend in end-to-end extraction performance across co-evolution rounds.

\clearpage
\clearpage
\clearpage

\clearpage
\section{Prompt Templates}
\label{sec:prompt-templates}

This appendix collects the shared task-prompt template and the module-specific prompts for optimization and regrouping. Placeholders are instantiated with the corresponding task context and program interfaces.

\subsection{Shared Task-Prompt Template of Agentic Module Optimization}
\label{sec:prompt-shared}

\begin{Verbatim}[fontsize=\footnotesize,breaklines=true,breakanywhere=false,breaksymbolleft={},breaksymbolright={},frame=none,baselinestretch=1]
TASK OBJECTIVE
{TASK_OBJECTIVE}

CONTEXT AVAILABLE
{CONTEXT_AVAILABLE}

PROGRAM MODIFICATION
Modify only the code between the #evolve-begin# and #evolve-end# markers in candidate.py. Preserve the following interfaces:
{PROGRAM_INTERFACE}

METRICS
{METRICS}

EXPERIMENT NOTEBOOK

Maintain notebook.json as a structured record of the local optimization process. Each entry corresponds to one candidate submission and is keyed by its candidate ID. Record the pre-submission analysis and post-evaluation review in the same entry.

You are responsible for writing the notebook. The controller independently archives the submitted program, evaluation metrics, execution results, and evidence files. Use these controller-generated records as the source of evaluation evidence; do not modify them.

1. BEFORE REVISING THE PROGRAM

Create an entry for the next candidate submission and record:
- Parent candidate: The ID of the evaluated candidate whose program you will modify.
- Problem: The specific program behavior to improve.
- Evidence: Evaluation results or examples demonstrating the problem, with references to the corresponding evidence files.
- Hypothesis: A testable explanation of the problem and why the proposed change should address it.
- Proposed change: The concrete modification to the candidate program.
- Expected outcome: Observable changes in metrics or example-level behavior that would support the hypothesis.

2. AFTER RECEIVING EVALUATION FEEDBACK

Update the same entry before starting another revision:
- Observations: The measured results and changes relative to the parent candidate, based on controller-generated feedback.
- Regressions: Previously successful cases or metrics that deteriorated. Explicitly state when no regressions were observed.
- Conclusion: Whether the hypothesis is supported, contradicted, or unresolved, with supporting evidence.
- Next action: The next experiment suggested by the results, or the reason for retaining the current candidate.
\end{Verbatim}

\subsection{Router Optimization}
\label{sec:prompt-router}

\begin{Verbatim}[fontsize=\footnotesize,breaklines=true,breakanywhere=false,breaksymbolleft={},breaksymbolright={},frame=none,baselinestretch=1]
TASK OBJECTIVE

Develop a router consisting of an index-construction method and a retrieval algorithm using the supplied training documents and their group labels. Optimize the router to meet the performance targets evaluated by the controller.

Optimize the router iteratively:
1. Read the experiment notebook and the latest evaluation feedback to understand previous attempts, unresolved problems, and regressions.
2. Inspect the original training documents and routing bad cases. Compare incorrectly ranked documents with examples from their assigned groups and the groups incorrectly ranked above them.
3. Follow the experiment-notebook procedure to record the problem, evidence, hypothesis, proposed change, and expected outcome.
4. Modify the permitted region of the candidate router program and invoke submit_candidate for independent evaluation by the controller.
5. Inspect the returned metrics and query-level results. Record the observations, regressions, conclusion, and next action in the same notebook entry.
6. Continue investigating and revising the router until the controller reports that its stopping condition has been met.

CONTEXT AVAILABLE

- Training documents: Original training documents available for inspecting document content and structure and developing the router.
  {TRAINING_DOCUMENTS_PATH}

- Current group assignments: The group label assigned to each training document, defining the routing targets.
  {PARENT_PARTITION_PATH}

- Evaluation splits: Fixed sub-training and held-out sub-validation splits of the training documents. For each split, the controller constructs the index using only the sub-training documents and evaluates routing using the held-out queries.
  {EVALUATION_SPLITS_PATH}

- Candidate router program: The editable router program, initialized from the selected parent version's router.
  {CANDIDATE_ROUTER_PATH}

- Reference router programs: Read-only router programs from the three highest-ranked historical versions, accompanied by their source-version metrics.
  {REFERENCE_ROUTER_PROGRAMS_PATH}

- Representation primitives: Callable OCR, BGE, and SigLIP functions available for use in the router program.
  {REPRESENTATION_PRIMITIVES_PATH}

- Experiment notebook: The structured record of the whole optimization process. Review previous attempts before each revision, and record the pre-submission analysis and post-evaluation review.
  {EXPERIMENT_NOTEBOOK_PATH}

- Evaluation feedback: The metrics and query-level bad cases of the current routing optimization. The end-to-end KIE extraction metrics and routing–extraction compatibility diagnostics of the base candidate version on the development set.
  {EVALUATION_FEEDBACK_PATH}

PROGRAM INTERFACE

Preserve both function signatures:
- build(training_view, config)
- query(document, resources)

METRICS:

METRICS FOR THE ROUTER OPTIMIZATION:
- Held-out Top-1 accuracy: The proportion of held-out queries whose top-ranked group matches the group label.

METRICS FOR THE END-TO-END KIE ON THE DEVELOPMENT SET:
- Field accuracy: The proportion of correct field predictions.
- Field F1: The F1 score for matching nonempty field predictions against references, averaged equally across target fields.
- Document accuracy: The proportion of documents with all target fields correctly extracted.
- Program coverage: The proportion of documents whose extraction-program outputs are accepted.
- Per-group program coverage: The program coverage of each group.
- Top-3 program acceptance rate: The proportion of documents accepted by at least one extraction program associated with their three highest-ranked groups.
- Per-group Top-3 rescue count: The number of documents whose routed Top-1 program rejects them but whose second- or third-ranked program accepts them, counted under the Top-1 group.
\end{Verbatim}

\subsection{Extraction Program Synthesis}
\label{sec:prompt-extraction}

\begin{Verbatim}[fontsize=\footnotesize,breaklines=true,breakanywhere=false,breaksymbolleft={},breaksymbolright={},frame=none,baselinestretch=1]
TASK OBJECTIVE

Develop a key information extraction program for the supplied training documents using the target field schema, extraction rules, and ground-truth annotations. The program must determine whether a document belongs to its supported scope and extract the target fields from accepted documents. Optimize extraction accuracy and in-group coverage while rejecting out-of-group documents, subject to the performance targets evaluated by the controller.

Optimize the extraction program iteratively:
1. Read the experiment notebook and the latest evaluation feedback to understand previous attempts, unresolved problems, and regressions.
2. Inspect the original training documents, reference annotations, and execution bad cases. Investigate incorrect field predictions, rejected in-group documents, and accepted out-of-group documents. Compare these cases with successfully processed documents to identify the causes.
3. Follow the experiment-notebook procedure to record the problem, evidence, hypothesis, proposed change, and expected outcome.
4. Modify the permitted region of the candidate extraction program and invoke submit_candidate for independent evaluation by the controller.
5. Inspect the returned metrics and record the observations, regressions, conclusion, and next action in the same notebook entry.
6. Continue investigating and revising the extraction program until the controller reports that its stopping condition has been met.

CONTEXT AVAILABLE

- In-group training documents: The target training documents to develop extraction rules and acceptance criteria.
  {IN_GROUP_TRAINING_DOCUMENTS_PATH}

- Out-of-group training documents: Training documents from other groups, including confusable examples, available for developing and evaluating rejection rules.
  {OUT_OF_GROUP_TRAINING_DOCUMENTS_PATH}

- Field annotations: Reference values for the target fields in the supplied in-group and out-of-group training documents.
  {FIELD_ANNOTATIONS_PATH}

- Field schema and extraction rules: The target field names, value formats, missing-value conventions, and field-specific extraction instructions.
  {FIELD_SCHEMA_AND_RULES_PATH}

- Candidate extraction program: The editable extraction program.
  {CANDIDATE_EXTRACTION_PROGRAM_PATH}

- Reference extraction programs: Read-only extraction programs from selected historical versions, accompanied by their source-version metrics and source-group information.
  {REFERENCE_EXTRACTION_PROGRAMS_PATH}

- Document-processing primitives: The callable OCR function available for use in the extraction program.
  {DOCUMENT_PROCESSING_PRIMITIVES_PATH}

- Experiment notebook: The structured record of the whole optimization process. Review previous attempts before each revision, and record the pre-submission analysis and post-evaluation review.
  {EXPERIMENT_NOTEBOOK_PATH}

- Evaluation feedback: The metrics, acceptance decisions, extracted fields, execution errors, and document-level bad cases of the current extraction optimization. Available historical feedback also includes end-to-end KIE metrics and group-specific extraction summaries on the development set.
  {EVALUATION_FEEDBACK_PATH}

PROGRAM INTERFACE

Preserve the function signature:
- extract(document)

RETURN:
- accept: A Boolean indicating whether the program accepts the document.
- fields: A mapping from the prescribed field names to extracted values, following the supplied field schema.

METRICS

METRICS FOR THE EXTRACTION PROGRAM OPTIMIZATION:
- Accepted in-group field F1: The F1 score for matching nonempty field predictions against references on accepted in-group documents, averaged equally across target fields.
- Accepted in-group document accuracy: The proportion of accepted in-group documents with all target fields correctly extracted.
- In-group coverage: The proportion of in-group documents accepted by the extraction program.
- Out-of-group false acceptance rate: The proportion of out-of-group documents accepted by the extraction program.

METRICS FOR THE END-TO-END KIE ON THE DEVELOPMENT SET:
- Field accuracy: The proportion of correct field predictions.
- Field F1: The F1 score for matching nonempty field predictions against references, averaged equally across target fields.
- Document accuracy: The proportion of documents with all target fields correctly extracted.
- Program coverage: The proportion of documents whose extraction-program outputs are accepted.
- Accepted-document field accuracy: Field accuracy on documents routed to the target group whose extraction-program outputs are adopted.
- Accepted-document field F1: Field F1 on the same accepted documents.
- Accepted-document document accuracy: Document accuracy on the same accepted documents.
- Per-group program coverage: The Program coverage of each group.
- Training–development field F1 gap: Field F1 on accepted in-group training documents minus field F1 on development documents routed to and accepted by the target extraction program.
- Training–development document accuracy gap: Document-level accuracy on accepted in-group training documents minus document-level accuracy on development documents routed to and accepted by the target extraction program.
\end{Verbatim}

\subsection{Routing-Guided Regrouping}
\label{sec:routing-regouping}

\begin{Verbatim}[fontsize=\footnotesize,breaklines=true,breakanywhere=false,breaksymbolleft={},breaksymbolright={},frame=none,baselinestretch=1]
TASK OBJECTIVE
Refine the current grouping using the optimized router's evaluation results and the failure cases. Identify incorrect assignments and unnecessarily separated groups, and propose document moves or group merges supported by the layout structural evidence.

Assess layout consistency based on the spatial arrangement of document elements, including headers, text blocks, tables, and key fields, and the applicable field extraction rules. Documents belong to the same layout group when these elements occupy similar relative positions, follow the same overall structure, and share the same field extraction rules, allowing minor differences in font, spacing, and scale.

Perform the review as follows
1. Read the current grouping and the optimized router's evaluation metrics and failure cases.
2. Inspect the documents involved in routing failures together with documents from their current groups and competing groups.
3. Compare their extraction-relevant organization. Determine whether the evidence supports a move or merge under the criteria.
4. Record supported changes and their document-level evidence in result.json. Preserve uncertain assignments and record unresolved issues in pending.
5. Check the proposal against the permitted operations and output schema, then invoke submit_candidate for controller validation.

CONTEXT AVAILABLE

- Training documents and parent grouping: Original training documents and their group assignments. Inspect these documents to compare information organization within and across groups.
  {TRAINING_DOCUMENTS_PATH}
  {PARENT_PARTITION_PATH}

- Evaluation splits: Fixed splits of the training set into sub-training sets for index construction and held-out sub-validation sets for evaluating the optimized router.
  {EVALUATION_SPLITS_PATH}

- Optimized router program: The router obtained from the preceding optimization step, available for inspecting its representation, index-construction, and retrieval logic.
  {OPTIMIZED_ROUTER_PROGRAM_PATH}

- Evaluation metrics and failure cases: Evaluation results for the optimized router on the held-out sub-validation sets defined above, including available query-level routing errors and competing group rankings.
  {ROUTER_EVALUATION_FEEDBACK_PATH}

PERMITTED CHANGES

MOVE CRITERION:
Move a document to another existing group when its spatial arrangement of headers, text blocks, tables, and key fields matches the destination group more closely than its current group. Identify the structural similarities with the destination group and differences from the current group, allowing minor variations in font, spacing, and scale.

MERGE CRITERION
Merge existing groups when their documents share the same overall structure and similar relative positions of headers, text blocks, tables, and key fields, allowing minor variations in font, spacing, and scale. In the output schema, a merge is represented by "moves" that reassign all documents from the source groups to one existing destination group.

EVIDENCE REQUIREMENTS
- Explain the relevant structural correspondences and differences.
- Allow superficial visual variation within a group.
- Treat routing errors and similarity scores as review signals; they alone do not justify a move or merge.
- Preserve the current assignment when the evidence is insufficient.
\end{Verbatim}

\begin{Verbatim}[fontsize=\footnotesize,breaklines=true,breakanywhere=false,breaksymbolleft={},breaksymbolright={},frame=none,baselinestretch=1]
OUTPUT SCHEMA

{
  "parent_version": "<supplied parent-version identifier>",
  "moves": [
    {
      "sample_id": "<training-document identifier>",
      "from_group": "<current group identifier>",
      "to_group": "<existing destination-group identifier>",
      "evidence": "<document-level or structural evidence>"
    }
  ],
  "retained": [
    {
      "sample_id": "<reviewed training-document identifier>",
      "evidence": "<evidence supporting the current assignment>"
    }
  ],
  "pending": [
    {
      "sample_id": "<unresolved training-document identifier>",
      "evidence": "<remaining uncertainty or missing evidence>"
    }
  ]
}
\end{Verbatim}

\subsection{Extraction-Guided Regrouping}
\label{sec:extraction-regouping}

\begin{Verbatim}[fontsize=\footnotesize,breaklines=true,breakanywhere=false,breaksymbolleft={},breaksymbolright={},frame=none,baselinestretch=1]
TASK OBJECTIVE

Refine the assigned training-document group using its extraction program, evaluation results, and original documents. Identify subsets whose extraction requirements are incompatible with the rules shared by the rest of the group, and propose evidence-supported splits.

Visual similarity alone does not establish extraction compatibility. Documents with similar layouts may require different rules for locating target fields, resolving semantic ambiguity, or parsing field values. Use execution feedback to investigate these differences and distinguish them from repairable defects in the current program.

Perform the review as follows
1. Read the extraction program and its evaluation metrics and failure cases, including incorrect predictions, rejected in-group documents, and falsely accepted out-of-group documents.
2. Inspect the corresponding original documents. Compare problematic in-group documents with successfully processed in-group documents, and use out-of-group documents to examine acceptance boundaries.
3. Trace recurring failures to document characteristics and the program rules affected by those characteristics.
4. Determine whether the evidence supports distinct extraction requirements under the split criterion below. Preserve the group when the failures can be addressed by correcting a shared rule.
5. Record supported subgroups, retained members, and supporting evidence in result.json. Preserve uncertain memberships and record unresolved issues in pending.
6. Check that the retained members and proposed subgroups cover the original group exactly once, then invoke submit_candidate for controller validation.

CONTEXT AVAILABLE

- In-group and out-of-group documents: Original documents assigned to the target group and confusable documents from other groups. Inspect in-group documents for differences in extraction requirements and use out-of-group documents as contrastive evidence.
  {IN_GROUP_DOCUMENTS_PATH}
  {OUT_OF_GROUP_DOCUMENTS_PATH}

- Extraction program: The current extraction program for the documents, including its field-extraction logic and acceptance criteria.
  {EXTRACTION_PROGRAM_PATH}

- Evaluation metrics and failure cases: Evaluation results for the extraction program, including available incorrect field predictions, acceptance and rejection errors, and execution failures.
  {EXTRACTION_EVALUATION_FEEDBACK_PATH}

PERMITTED CHANGES

A. SPLIT CRITERION
Split the current group when document-level and execution evidence identify subsets requiring different rules for field localization, semantic disambiguation, or value parsing.

For each proposed subgroup:
- Identify observable document characteristics defining its membership.
- Explain which extraction rules differ from those used for the remaining documents and why.
- Cite execution cases demonstrating the consequences of applying the current shared rules.
- Explain why the difference reflects distinct extraction rules rather than a fixable defect in the program's implementation.

B. EVIDENCE REQUIREMENTS
- Similar-looking documents may be separated when their extraction requirements differ.
- A single extraction failure, OCR noise, or a missing field alone does not justify a split.
- Out-of-group false acceptance can reveal weaknesses in acceptance criteria, but alone does not justify splitting the current group.
- Define subgroup membership through document characteristics rather than document identifiers or current program success.

METRICS

- Accepted in-group field F1: The F1 score for matching nonempty field predictions against references on accepted in-group documents, averaged equally across target fields.
- Accepted in-group document accuracy: The proportion of accepted in-group documents with all target fields correctly extracted.
- In-group coverage: The proportion of in-group documents accepted by the extraction program.
- Out-of-group false acceptance rate: The proportion of out-of-group documents accepted by the extraction program.
\end{Verbatim}

\begin{Verbatim}[fontsize=\footnotesize,breaklines=true,breakanywhere=false,breaksymbolleft={},breaksymbolright={},frame=none,baselinestretch=1]
OUTPUT SCHEMA

{
  "parent_version": "<supplied parent-version identifier>",
  "group_id": "<current group identifier>",
  "retained_members": [
    "<training-document identifier retained in the original group>"
  ],
  "new_groups": [
    {
      "temporary_id": "<unique temporary subgroup identifier>",
      "sample_ids": [
        "<training-document identifier assigned to this subgroup>"
      ],
      "evidence": "<structural evidence supporting this subgroup>"
    }
  ],
  "evidence": "<overall justification for the proposed split>",
  "pending": []
}
\end{Verbatim}

\end{document}